\documentclass[11pt]{article}
\usepackage[dvips]{graphicx}
\usepackage{textcomp}
\usepackage{amsbsy}
\usepackage{latexsym}
\usepackage[mathscr]{eucal}
\usepackage{amsfonts,amsmath,amsthm}
\usepackage{amssymb}
\usepackage[usenames]{xcolor}

\usepackage{fullpage}

\begin{document}
\bibliographystyle{plain}
\title
{Limitations  on approximation by  deep and shallow neural networks 
}
\author{ 
Guergana Petrova and  Przemys{\l}aw Wojtaszczyk
\thanks{%
   G.P.  was supported by the  NSF Grant  DMS 2134077, Tripods Grant CCF-1934904, and ONR Contract N00014-20-1-278. 
     }}
\hbadness=10000
\vbadness=10000
\newtheorem{lemma}{Lemma}[section]
\newtheorem{prop}[lemma]{Proposition}
\newtheorem{cor}[lemma]{Corollary}
\newtheorem{theorem}[lemma]{Theorem}
\newtheorem{remark}[lemma]{Remark}
\newtheorem{example}[lemma]{Example}
\newtheorem{definition}[lemma]{Definition}
\newtheorem{proper}[lemma]{Properties}
\newtheorem{assumption}[lemma]{Assumption}
%
\newenvironment{disarray}{\everymath{\displaystyle\everymath{}}\array}{\endarray}

\def\RR{\rm \hbox{I\kern-.2em\hbox{R}}}
\def\NN{\rm \hbox{I\kern-.2em\hbox{N}}}
\def\ZZ{\rm {{\rm Z}\kern-.28em{\rm Z}}}
\def\CC{\rm \hbox{C\kern -.5em {\raise .32ex \hbox{$\scriptscriptstyle
|$}}\kern
-.22em{\raise .6ex \hbox{$\scriptscriptstyle |$}}\kern .4em}}
\def\vp{\varphi}
\def\<{\langle}
\def\>{\rangle}
\def\t{\tilde}
\def\i{\infty}
\def\e{\varepsilon}
\def\sm{\setminus}
\def\nl{\newline}
\def\o{\overline}
\def\wt{\widetilde}
\def\wh{\widehat}
\def\cT{{\cal T}}
\def\cA{{\cal A}}
\def\cI{{\cal I}}
\def\cV{{\cal V}}
\def\cB{{\cal B}}
\def\cF{{\cal F}}
\def\cY{{\cal Y}}

\def\cD{{\cal D}}
\def\cP{{\cal P}}
\def\cJ{{\cal J}}
\def\cM{{\cal M}}
\def\cO{{\cal O}}
\def\Chi{\raise .3ex
\hbox{\large $\chi$}} \def\vp{\varphi}
\def\lsima{\hbox{\kern -.6em\raisebox{-1ex}{$~\stackrel{\textstyle<}{\sim}~$}}\kern -.4em}
\def\lsim{\hbox{\kern -.2em\raisebox{-1ex}{$~\stackrel{\textstyle<}{\sim}~$}}\kern -.2em}
\def\[{\Bigl [}
\def\]{\Bigr ]}
\def\({\Bigl (}
\def\){\Bigr )}
\def\[{\Bigl [}
\def\]{\Bigr ]}
\def\({\Bigl (}
\def\){\Bigr )}
\def\L{\pounds}
\def\pr{{\rm Prob}}
\newcommand{\cs}[1]{{\color{magenta}{#1}}}
\def\ds{\displaystyle}
\def\ev#1{\vec{#1}}     
\newcommand{\lt}{\ell^{2}(\nabla)}
\def\Supp#1{{\rm supp\,}{#1}}
\def\R{\mathbb{R}}
\def\E{\mathbb{E}}
\def\nl{\newline}
\def\T{{\relax\ifmmode I\!\!\hspace{-1pt}T\else$I\!\!\hspace{-1pt}T$\fi}}
\def\N{\mathbb{N}}
\def\Z{\mathbb{Z}}
\def\N{\mathbb{N}}
\def\Zd{\Z^d}
\def\Q{\mathbb{Q}}
\def\C{\mathbb{C}}
\def\Rd{\R^d}
\def\gsim{\mathrel{\raisebox{-4pt}{$\stackrel{\textstyle>}{\sim}$}}}
\def\sime{\raisebox{0ex}{$~\stackrel{\textstyle\sim}{=}~$}}
\def\lsim{\raisebox{-1ex}{$~\stackrel{\textstyle<}{\sim}~$}}
\def\div{\mbox{ div }}
\def\M{M}  \def\NN{N}                  
\def\Le{{\ell^1}}            
\def\Lz{{\ell^2}}
\def\Let{{\tilde\ell^1}}     
\def\Lzt{{\tilde\ell^2}}
\def\Ltw{\ell^\tau^w(\nabla)}
\def\t#1{\tilde{#1}}
\def\la{\lambda}
\def\La{\Lambda}
\def\ga{\gamma}
\def\BV{{\rm BV}}
\def\Ga{\eta}
\def\al{\alpha}
\def\cZ{{\cal Z}}
\def\cA{{\cal A}}
\def\cU{{\cal U}}
\def\argmin{\mathop{\rm argmin}}
\def\argmax{\mathop{\rm argmax}}
\def\prob{\mathop{\rm prob}}

\def\cO{{\cal O}}
\def\cA{{\cal A}}
\def\cC{{\cal C}}
\def\cS{{\cal F}}
\def\bu{{\bf u}}
\def\bz{{\bf z}}
\def\bZ{{\bf Z}}
\def\bI{{\bf I}}
\def\cE{{\cal E}}
\def\cD{{\cal D}}
\def\cG{{\cal G}}
\def\cI{{\cal I}}
\def\cJ{{\cal J}}
\def\cM{{\cal M}}
\def\cN{{\cal N}}
\def\cT{{\cal T}}
\def\cU{{\cal U}}
\def\cV{{\cal V}}
\def\cW{{\cal W}}
\def\cL{{\cal L}}
\def\cB{{\cal B}}
\def\cG{{\cal G}}
\def\cK{{\cal K}}
\def\cX{{\cal X}}
\def\cS{{\cal S}}
\def\cP{{\cal P}}
\def\cQ{{\cal Q}}
\def\cR{{\cal R}}
\def\cU{{\cal U}}
\def\bL{{\bf L}}
\def\bl{{\bf l}}
\def\bK{{\bf K}}
\def\bC{{\bf C}}
\def\X{X\in\{L,R\}}
\def\ph{{\varphi}}
\def\D{{\Delta}}
\def\H{{\cal H}}
\def\bM{{\bf M}}
\def\bx{{\bf x}}
\def\bj{{\bf j}}
\def\bG{{\bf G}}
\def\bP{{\bf P}}
\def\bW{{\bf W}}
\def\bT{{\bf T}}
\def\bV{{\bf V}}
\def\bv{{\bf v}}
\def\bt{{\bf t}}
\def\bz{{\bf z}}
\def\bw{{\bf w}}
\def \span{{\rm span}}
\def \meas {{\rm meas}}
\def\rhom{{\rho^m}}
\def\diff{\hbox{\tiny $\Delta$}}
\def\EE{{\rm Exp}}
\def\lll{\langle}
\def\argmin{\mathop{\rm argmin}}
\def\codim{\mathop{\rm codim}}
\def\rank{\mathop{\rm rank}}

\def\argmax{\mathop{\rm argmax}}
\def\dJ{\nabla}
\newcommand{\ba}{{\bf a}}
\newcommand{\bb}{{\bf b}}
\newcommand{\bc}{{\bf c}}
\newcommand{\bd}{{\bf d}}
\newcommand{\bs}{{\bf s}}
\newcommand{\bff}{{\bf f}}
\newcommand{\bp}{{\bf p}}
\newcommand{\bg}{{\bf g}}
\newcommand{\by}{{\bf y}}
\newcommand{\br}{{\bf r}}
\newcommand{\be}{\begin{equation}}
\newcommand{\ee}{\end{equation}}
\newcommand{\bea}{$$ \begin{array}{lll}}
\newcommand{\eea}{\end{array} $$}
\def \Vol{\mathop{\rm  Vol}}
\def \mes{\mathop{\rm mes}}
\def \Prob{\mathop{\rm  Prob}}
\def \exp{\mathop{\rm    exp}}
\def \sign{\mathop{\rm   sign}}
\def \sp{\mathop{\rm   span}}
\def \rad{\mathop{\rm   rad}}
\def \vphi{{\varphi}}
\def \csp{\overline \mathop{\rm   span}}
%
%
\newcommand{\beqn}{\begin{equation}}
\newcommand{\eeqn}{\end{equation}}
\def\beginproof{\noindent{\bf Proof:}~ }
\def\endproof{\hfill\rule{1.5mm}{1.5mm}\\[2mm]}

\newenvironment{Proof}{\noindent{\bf Proof:}\quad}{\endproof}

\renewcommand{\theequation}{\thesection.\arabic{equation}}
\renewcommand{\thefigure}{\thesection.\arabic{figure}}

\makeatletter
\@addtoreset{equation}{section}
\makeatother

\newcommand\abs[1]{\left|#1\right|}
\newcommand\clos{\mathop{\rm clos}\nolimits}
\newcommand\trunc{\mathop{\rm trunc}\nolimits}
\renewcommand\d{d}
\newcommand\dd{d}
\newcommand\diag{\mathop{\rm diag}}
\newcommand\dist{\mathop{\rm dist}}
\newcommand\diam{\mathop{\rm diam}}
\newcommand\cond{\mathop{\rm cond}\nolimits}
\newcommand\eref[1]{{\rm (\ref{#1})}}
\newcommand{\iref}[1]{{\rm (\ref{#1})}}
\newcommand\Hnorm[1]{\norm{#1}_{H^s([0,1])}}
\def\int{\intop\limits}
\renewcommand\labelenumi{(\roman{enumi})}
\newcommand\lnorm[1]{\norm{#1}_{\ell^2(\Z)}}
\newcommand\Lnorm[1]{\norm{#1}_{L_2([0,1])}}
\newcommand\LR{{L_2(\R)}}
\newcommand\LRnorm[1]{\norm{#1}_\LR}
\newcommand\Matrix[2]{\hphantom{#1}_#2#1}
\newcommand\norm[1]{\left\|#1\right\|}
\newcommand\ogauss[1]{\left\lceil#1\right\rceil}
\newcommand{\QED}{\hfill
\raisebox{-2pt}{\rule{5.6pt}{8pt}\rule{4pt}{0pt}}%
  \smallskip\par}
\newcommand\Rscalar[1]{\scalar{#1}_\R}
\newcommand\scalar[1]{\left(#1\right)}
\newcommand\Scalar[1]{\scalar{#1}_{[0,1]}}
\newcommand\Span{\mathop{\rm span}}
\newcommand\supp{\mathop{\rm supp}}
\newcommand\ugauss[1]{\left\lfloor#1\right\rfloor}
\newcommand\with{\, : \,}
\newcommand\Null{{\bf 0}}
\newcommand\bA{{\bf A}}
\newcommand\bB{{\bf B}}
\newcommand\bR{{\bf R}}
\newcommand\bD{{\bf D}}
\newcommand\bE{{\bf E}}
\newcommand\bF{{\bf F}}
\newcommand\bH{{\bf H}}
\newcommand\bU{{\bf U}}
\newcommand \A {{\bb A}}
\newcommand\cH{{\cal H}}
\newcommand\sinc{{\rm sinc}}
\def\enorm#1{| \! | \! | #1 | \! | \! |}

\newcommand{\am}{a_{\min}}
\newcommand{\aM}{a_{\max}}

\newcommand{\dm}{\frac{d-1}{d}}

\let\bm\bf
\newcommand{\bbeta}{{\mbox{\boldmath$\beta$}}}
\newcommand{\bal}{{\mbox{\boldmath$\alpha$}}}
\newcommand{\bbi}{{\bm i}}

\def\nnew{\color{Red}}
\def\onew{\color{orange}}
\def\wnew{\color{magenta}}

\newcommand{\dI}{\Delta}
\newcommand\aconv{\mathop{\rm absconv}}

\maketitle

\vskip 0.2in

  \begin{abstract}   
  
  We prove Carl's type inequalities for the error of  approximation of compact sets $\cK$ by 
  deep and shallow neural networks.   This in turn gives lower bounds on how well we can approximate
  the functions in $\cK$ when requiring the approximants to come from outputs of such networks.
    Our  results are obtained as a byproduct  of 
  the study of the recently introduced Lipschitz widths.    
  
  \noindent
  {\bf AMS subject classification:} 41A46, 41A65, 82C32
  
  \noindent
  {\bf Key words:} widths, entropy numbers, neural networks
        \end{abstract}

\section{Introduction}
 Since neural network approximation is the method of choice in building numerical algorithms in many application areas, it is important to understand not only how well they approximate but also any lower bounds on their approximation power.  In this paper, we  study  the  limitations of  deep and shallow neural networks  to approximate  a  compact subset $\mathcal K\subset X$ of a Banach space $X$ when it is required that the parameters in the approximation procedure have certain bounds.
 This is done by  proving appropriate Carl's type inequalities that relate the error of neural network approximation of  $\mathcal K$ to the entropy numbers of this set.

We consider  feed-forward  neural networks (NN) with  ReLU or Lipschitz sigmoidal activation functions,  width 
$W\geq 2$ and depth $n$,  whose parameters have absolute values bounded by  a given function $w(n)$. We 
 prove that  the  capabilities  of these networks   to approximate any compact subset $\mathcal K$ is limited by the behavior of its
 entropy numbers. 
 For example, we show that  Deep Neural Networks (DNNs)  with constant width $W$, depth $n$ and parameters bounded by $w(n)=Cn^\delta$,  $\delta> 0$,  cannot approximate better than 
 $C[\log_2n]^{\beta-\alpha}n^{-2\alpha}$ any compact set of functions $\mathcal K$ whose entropy numbers $\epsilon_n(\mathcal K)_X\gtrsim [\log_2n]^{\beta}n^{-\alpha}$, $n=1,2,\ldots$, see Corollary \ref{mainc}. 
 We also   show that  if a  class of functions $\mathcal K$ is approximated up to accuracy
  $C[\log_2n]^{\beta}n^{-\alpha}$, $n=1,2, \ldots,$ by the above mentioned  DNNs, then  $\epsilon_n(\mathcal K)_X\leq C'n^{-\frac{\alpha}{2}}[\log_2n]^{\beta+\frac{\alpha}{2}}$, see Corollary \ref{NNA}. 
    In particular, we obtain estimates for the  entropy numbers of the classes  of functions 
 that are approximated via DNNs with predetermined rates (approximation classes)  as the depth $n$ of the NN grows. 
 
  Results of  the same form  are obtained   for shallow NNs (NNs with one hidden layer) as we let the 
  width $W\to \infty$.  
  For example, we show that  SNNs  with width $W$ and parameters bounded by $w(W)=CW^\delta$,  $\delta\geq  0$,  cannot approximate better than 
 $C[\log_2W]^{\beta-\alpha}W^{-\alpha}$ a compact set of functions $\mathcal K$ whose entropy numbers are $\gtrsim [\log_2n]^{\beta}n^{-\alpha}$, see Corollary \ref{maincs}. 
 Also, if a  class of functions $\mathcal K$ is approximated up to accuracy
  $C[\log_2W]^{\beta}W^{-\alpha}$, $W=2,3, \ldots,$ by a SNN, then  it has    entropy numbers $\epsilon_n(\mathcal K)_X\leq C'n^{-\alpha}[\log_2n]^{\beta+\alpha}$,  $n\in \mathbb N$, see Corollary \ref{NNAs}. 
Analogous estimates for general  bounds $w(n)$, respectively $w(W)$,  on the parameters of deep and shallow NNs are also given, including the case $w(n)=C2^{cn^\nu}$.
  
 The novelty of our results is that the lower bounds we prove apply to any compact set $\cK$ and any Banach space
 norm $X$. Previous results have proven lower bounds on approximation by neural networks using the VC dimension of neural network spaces.  However, such results have been shown only for very specific model classes (see the discussion in the conclusion section).  
 
 In our analysis of NN approximation, we are not concerned with the numerical aspect of the construction of the corresponding DNN and its stability, but rather with the theoretical lower 
 bounds of the performance of such an approximation. We show that the mapping that assigns to each choice of NN parameters a function, generated by the NN feed-forward architecture, 
  is a Lipschitz mapping with a large Lipschitz constant, depending of the upper bound $w(n)$ 
  (or $w(W)$ in the case of shallow NN) of the NN parameters. Thus, we can view 
 NN approximation as  an approximation of a class $\mathcal K$ via Lipschitz mappings.  This  type of approximation is studied via the introduced in \cite{PW} Lipschitz widths $d_n^\gamma(\mathcal K)_X$.
  These widths  join the  plethora of classical widths available, see \cite{DKLT}, and give a theoretical bound on the approximation of $\mathcal K$ via $\gamma$-Lipschitz mappings defined on unit balls in 
  $\mathbb R^n$.  An almost complete analysis of these widths with parameter $\gamma=const$ or $\gamma=\gamma_n=\lambda^n$, $\lambda>1$,  was given in \cite{PW}. 
 
Note that  DNNs whose  parameters are  bounded by  $w(n)$  are  Lipschitz mappings and are associated with Lipschitz widths with constant  $\gamma=\gamma_n$, where 
$2^{c_1n(1+\log_2w(n))}< \gamma_n< 2^{c_2n(1+\log_2w(n))}$, see Theorem \ref{NN} and \eref{main}.
 Shallow NNs  (SNNs) whose  parameters are  bounded by  $w(W)$  are Lipschitz mappings associated with Lipschitz widths with constant $\gamma=\gamma_W=2^{c(\log_2W+\log_2w(n))}$, 
 see Theorem \ref{NNs} and \eref{mains}.
When  $w(n)=const$ we have  $\gamma=\gamma_n=\lambda^n$ and such DNNs  have been studied in \cite{PW} via the corresponding Lipschitz widths. The
  investigation of  the approximation power  of  deep or shallow NNs with a general bound $w(n)$ of  their parameters requires  a study of Lipschitz widths with  
  Lipschitz constant  $\gamma=\gamma_n=2^{\varphi(n)}$ with 
  rather general functions $\varphi$. In this paper, we provide such a study and its consequences for NN approximation. 

 Carl's type inequalities for the approximation power of SNNs have been derived in \cite{JS} for compact sets $\mathcal K$ that are the closures of the symmetric convex 
hull of certain dictionaries and NNs with bounded parameters and certain activation functions.  Lower bounds for DNN approximation  
for sets $\mathcal K$ that are the unit ball of certain Besov classes have been obtained in \cite{DHP}, see \S 5.9 and the references therein. These estimates rely on  bounds on 
 the VC dimension of DNNs and the particular structure of the sets $\mathcal K$. The approach proposed in this paper uses only the behavior of the entropy numbers of a general  compact 
 set $\mathcal K$ and thus can be applied to various novel classes.
 
  The paper is organized as follows. In \S \ref{S2},  we introduce our notation and recall the definitions of NNs, Lipschitz widths and entropy numbers. 
 We show in \S\ref{DNNLM} that the NNs under consideration are Lipschitz mappings. We revisit the definition of Lipschitz width in \S\ref{S3}. 
 Lower bounds for the Lipschitz widths $d_n^{\gamma_n}(\mathcal K)_X$ with Lipschitz constants $\gamma_n=2^{\varphi(n)}$ for a compact class $\mathcal K$ and their implication for 
 deep and shallow NN approximation of $\mathcal K$ are provided in \S\ref{LB}.  Upper bounds for the entropy numbers of a class $\mathcal K$ that use the behavior of the
 Lipschitz widths and the  NN approximation rates for $\mathcal K$
   are presented in \S\ref{UB}. Further properties  of $d_n^{\gamma_n}(\mathcal K)_X$ are discussed in \S\ref{p1} and \S\ref{p2a}. 
 Finally, our concluding remarks are presented in \S\ref{con} and some lemmas and their proofs are discussed in \S\ref{ap}.
 
 
\section{Preliminaries}
\label{S2}

 In this section, we introduce our notation and recall some known facts about NNs, Lipschitz widths and entropy numbers. In what follows, we will denote by  
 $A\gtrsim B$ the fact that there is an absolute constant $c>0$ such that $A\geq cB$, where $A,B$ are some expressions that depend on $n$ as $n\to\infty$. 
 Note that  the value of $c$ may change from line to line, but is always independent on $n$. Similarly, we 
 will use the notation $A\lesssim B$, which is  defined in an analogues way,  and $A\asymp B$ if $A\gtrsim B$ and $A\lesssim B$.

\subsection{Neural networks}
\subsubsection{Deep feed-forward neural networks}
 A feed-forward NN with activation function 
$\sigma:\mathbb R\to\mathbb R$, constant width $W$ and depth $n$ generates a family $\Sigma_{n,\sigma}$ of continuous functions  
$$
\Sigma_{n,\sigma}:=\{\Phi_\sigma(y): \,\,y\in \mathbb R^{\tilde n},\,\,\tilde n=\tilde n(W,n)=C_0(W)n\} \subset C(\Omega), \quad \Omega\subset \mathbb R^d,
$$ 
that is used to produce an approximant to a given function $f\in \mathcal K$ or the whole class $\mathcal K$. 
Each parameter vector $y\in \mathbb R^{\tilde n}$ determines a continuous function $\Phi_\sigma(y)\in \Sigma_{n,\sigma}$, defined on $\Omega$,  of the form
\begin{equation}
\label{NN1}
\Phi_\sigma(y):=A^{(n)}\circ\bar\sigma\circ A^{(n-1)}\circ\ldots\circ \bar\sigma\circ A^{(0)},
\end{equation}
where  $\bar\sigma:\mathbb R^W\to\mathbb R^W$ is given  by 
\begin{equation}
\label{bars}
\bar\sigma(z_{j+1},\ldots,z_{j+W})=(\sigma(z_{j+1}),\ldots,\sigma(z_{j+W})),
\end{equation}
and  $A^{(0)}:\mathbb R^d\to\mathbb R^W$, $A^{(\ell)}:\mathbb R^{W}\to\mathbb R^{W}$, $\ell=1,\ldots,n-1$,  and $A^{(n)}:\mathbb R^W\to\mathbb R$ are affine mappings.
Note that $y\in\mathbb R^{\tilde n}$  is the vector with coordinates  the entries of the matrices and offset   vectors (biases) of the affine mappings $A^{(\ell)}$,
$\ell=0,\ldots,n$. We order them   in such a way that the entries of $A^{(\ell)}$
 appear before those of $A^{(\ell+1)}$ and the ordering  for each $A^{(\ell)}$ is done in  the same way. For detailed study of such DNNs we refer the reader to \cite{DHP} and the references therein.
 We investigate the approximation power of $\Sigma_{n,\sigma}$ when the width $W$ is fixed and  the depth  $n\to \infty$.

\subsubsection{Shallow neural networks}
 A shallow NN with activation function 
$\sigma:\mathbb R\to\mathbb R$, and width $W$  generates a family $\Xi_{\sigma,W}$ of continuous functions  
$$
\Xi_{W,\sigma}:=\{\Psi_\sigma(y): \,\,y\in \mathbb R^{\tilde W},\,\,\tilde W=C_0(d)W\} \subset C(\Omega), \quad \Omega\subset \mathbb R^d,
$$ 
that is used to produce an approximant to a given function $f\in \mathcal K$ or the whole class $\mathcal K$. 
Each parameter vector $y\in \mathbb R^{\tilde W}$ determines a continuous function $\Psi_\sigma(y)\in \Xi_{W,\sigma}$, defined on $\Omega$,  of the form
\begin{equation}
\label{NNsdef}
\Psi_\sigma(y):=A^{(1)}\circ\bar\sigma\circ  A^{(0)},
\end{equation}
where  $\bar\sigma:\mathbb R^W\to\mathbb R^W$ is given  by 
\begin{equation}
\label{bars}
\bar\sigma(z_{j+1},\ldots,z_{j+W})=(\sigma(z_{j+1}),\ldots,\sigma(z_{j+W})),
\end{equation}
and  $A^{(0)}:\mathbb R^d\to\mathbb R^W$  and $A^{(1)}:\mathbb R^W\to\mathbb R$ are affine mappings. We investigate the approximation power of $\Xi_{W,\sigma}$ as the width $W\to \infty$.

\subsection{Lipschitz widths}
Lipschitz widths $d_n^\gamma({\mathcal K})_X$ for a compact subset  $\mathcal K\subset X$ of a Banach space $X$ with a Lipschitz constant $\gamma=C_0=const$ or $\gamma=\gamma_n=C'\lambda^n$ with $\lambda>1$ were introduced and analyzed in \cite{PW}.
The latter  were used to study lower bounds for ReLU DNNs with weights and biases bounded by $1$. However, in practice,  the weights and biases used in a DNN may grow. This growth affects the Lipschitz constant associated with the 
corresponding DNN  viewed as a Lipschitz mapping. Thus, providing lower bounds for the approximation power of such networks requires  the investigation of Lipschitz widths with various Lipschitz constants 
$\gamma$ that depend on $n$.

Let us first  recall the definition of $d_n^\gamma({\mathcal K})_X$. 
We denote by    $({\mathbb R}^n,\|.\|_{Y_n})$, $n\geq 1$,
the $n$-dimensional Banach space with a fixed norm $\|\cdot\|_{Y_n}$, by
$$
 B_{Y_n}(r):=\{y\in {\mathbb R}^n:\,\,\|y\|_{Y_n}\leq r\},
 $$
its  ball with radius $r$, and by
$$
\|y\|_{\ell_\infty^n}:=\max_{j=1,\ldots,n}|y_j|, \quad \|y\|_{\ell_p^n}:=\Big(\sum_{j=1,\ldots,n}|y_j|^p\Big )^{1/p}, \quad 1\leq p<\infty,
$$
the usual $\ell_p$ norms of  $y=(y_1,\ldots,y_n)\in \mathbb R^n$.
For  $\gamma\geq 0$, 
 the {\it fixed Lipschitz} width $d^\gamma({\mathcal K},Y_n)_X$ is defined as
\begin{equation}
\label{L1}
 d^0({\mathcal K},Y_n)_X:={\rm rad}(\cK):=\inf_{g\in X}\sup_{f\in \cK}\|g-f\|_X,\quad d^\gamma({\mathcal K},Y_n)_X:= \inf_{{\mathcal L}_n} \sup_{f\in {\mathcal K}}\inf_{y\in B_{Y_n}(1)}\|f-{\mathcal L}_n(y)\|_X,
\end{equation}
where the infimum is taken over all Lipschitz mappings
 $$
 {\mathcal L}_n:(B_{Y_n}(1),\|\cdot\|_{Y_n})\to X,
 $$ 
that satisfy  
 the Lipschitz condition 
\begin{equation}
\label{L2}
 \sup_{y,y'\in B_{Y_n}(1)}
 \frac{\|{\mathcal L}_n(y)-{\mathcal L}_n(y')\|_X}{\|y-y'\|_{Y_n}}\leq \gamma,
\end{equation}
 with constant $\gamma$. 
Then, the
{\it Lipschitz} width $d_n^\gamma({\mathcal K})_X$ is
\begin{equation}
\nonumber
 d_n^0({\mathcal K})_X:=d^0({\mathcal K},Y_n)_X, \quad    d_n^\gamma({\mathcal K})_X:=\inf_{\|\cdot\|_{Y_n}}d^\gamma({\mathcal K},Y_n)_X,
\end{equation}
where the infimum is taken over  all norms $\|\cdot\|_{Y_n}$ in ${\mathbb R}^n$.    

Various notions of widths had been introduced and used in 
approximation theory to theoretically quantify the limitations of certain types of approximations. We refer the reader to \cite{DKLT} or \cite{LGM}, where 
different widths and their decay rates for common smoothness classes have been discussed. 
Note that   the definition of Lipschitz width  is similar to the definition of the manifold $n$-width  $\delta_n(\mathcal K)_X$,  (see e.g. \cite{CDPW})
$$
\delta_n(\mathcal K)_X:={\rm inf}_{M,a}{\rm sup}_{f\in \mathcal K}\|f-M(a(f))\|_X,
$$
where the infimum is taken over all continuous mappings $a:\mathcal K\to \mathbb R^n$, $M:\mathbb R^n\to X$. However, in the definition of Lipschitz width, 
we  impose the  stronger Lipschitz condition on the approximation mapping.

Before going further, we list some of the properties of the Lipschitz width $d_n^\gamma(\mathcal K)_X$, proved in \cite{PW}, which we  gather  in the following theorem.
\begin{theorem}
\label{specialnorm}
For any   $n\in \mathbb N$, any compact set $\mathcal K\subset X$,  and any constant $\gamma>0$ we have:
\begin{itemize}
\item  $d_n^{\gamma}(\mathcal K)_X$  is a monotone decreasing function of $\gamma$ and $n$. More precisely,
\begin{itemize}
	\item If $\gamma_1\leq \gamma_2$ then $d_n^{\gamma_2}(\mathcal K)_X\leq d_n^{\gamma_1}(\mathcal K)_X$;
       \item If $n_1\leq n_2$ then $d_{n_2}^{\gamma}(\mathcal K)_X\leq d_{n_1}^{\gamma}(\mathcal K)_X$.
 \end{itemize}
\item there is  a 
norm $\|\cdot\|_{\mathcal Y_n}$ on $\mathbb R^n$  such that for every $y\in \mathbb R^n$ we have  $\|y\|_{\ell_\infty^n}\leq \|y\|_{\mathcal Y_n}\leq \|y\|_{\ell_1^n}$,
and
$$
d_n^\gamma(\mathcal K)_X=d^\gamma(\mathcal K,{\mathcal Y}_n)_X.
$$
\end{itemize}
\end{theorem}

\subsection{Entropy numbers}

 We recall, see  e.g. \cite{C, CS, LGM}, that
the {\it entropy numbers} $\epsilon_n({\mathcal K}) _X$, $n\geq 0$, of a compact set ${\mathcal K}\subset X$ are defined 
as the infimum of all $\epsilon>0$ for which $2^n$ balls with centers from $X$ and radius $\epsilon$ cover ${\mathcal K}$.  
Formally,  we  write
$$ 
\epsilon_n({\mathcal K})_X=\inf\{ \epsilon>0 \ :\ {\mathcal K} 
\subset \bigcup_{j=1}^{2^n} B(g_j,\epsilon), \ g_j\in X, \ j=1,\ldots,2^n\}.
$$


\section{Neural networks are Lipschitz mappings}
\label{DNNLM}
Our choice of norm when 
working with NNs is the  $\|\cdot\|_{\ell_\infty^{ n}}$ norm of the  parameters $y$ of  the neural network. This is simply because 
 we are interested in the asymptotic behavior (with  respect to the depth $n$ of the  DNN or the width $W$ of the SNN) 
 of the approximation error that the network provides for a a class $\cK$ and not in the best possible constants
 in the error estimates.

\subsection{Deep neural networks}
 
We do not  investigate 
 what information about the function $f$ is given or what methods one employs to 
find an appropriate parameter vector $y^*\in \mathbb R^{\tilde n}$ such that the function $\Phi(y^*)$ is the (near)best approximant to $f$ from the set $\Sigma_{n,\sigma}$, but rather focus 
on  the properties of the mapping 
$$
y\to \Phi_\sigma(y), \quad \Phi_\sigma(y)\in \Sigma_{n,\sigma},
$$
where all parameters (entries of the matrices and biases) are bounded by $w(n)$, where $w(n)\geq 1$.
{ To illustrate  the dependence on $w(n)$, we denote the collection of all such mappings as $\Sigma_{n,\sigma}(w(n))$,
namely
$$
\Sigma_{n,\sigma}(w(n)):=\Phi_\sigma(B_{\ell_\infty^{\tilde n}}(w(n)),
$$
with $\Phi_\sigma$ being defined in (\ref{NN1}).}
 We have shown in \cite{PW}   that    in the case $w(n)\equiv 1$, $\sigma={\rm ReLU}$, and $\Omega=[0,1]^d$, the mapping
$$
\Phi_\sigma:(B_{\ell_\infty^{\tilde n}}(1),\|\cdot\|_{\ell_\infty^{\tilde n}})\to C(\Omega), \quad  \tilde n=C_0n, \quad C_0=C_0(W),
$$
defined in (\ref{NN1}) is a Lipschitz mapping with Lipschitz constant $L_n$,  where $2^{c_2n}<L_n<2^{c_1n}$ for fixed constants $c_1,c_2>0$ 
depending on the width $W$. More precisely, we have
$$
\|\Phi_\sigma(z)-\Phi_\sigma(y)\|_{C([0,1]^d)}\leq L_n\|z-y\|_{\ell_\infty^{\tilde n}}\quad \hbox{for all}\quad z,y\in B_{\ell_\infty^{\tilde n}}(1).
$$
 Here, we will investigate what is the  Lipschitz constant  $L_n$ when the parameters $y$ have components bounded by $w(n)$, namely
 $y\in B_{\ell_\infty^{\tilde n}}(w(n))$ and the activation function is ${\rm ReLU}(t):=\max\{0,t\}=t_+$ or  is a
 Lipschitz sigmoidal function $\sigma$ with Lipschitz constant $L$. The properties that we are going to use about the sigmoidal function $\sigma$ are 
 $$
 |\sigma(t)|\leq 1, \quad |\sigma(t_1)-\sigma(t_2)|\leq L|t_1-t_2|, \quad t_1,t_2\in \mathbb R.
 $$
 Clearly, for any $m$, vectors  $ \bar y,\hat y,\eta \in \mathbb R^m$ and  numbers $y_0,\hat y_0\in\mathbb R$, where $\bar y,y_0$ and $\hat y,\hat y_0$  are  subsets  of the coordinates  of $y$ and $y'$, respectively, we have
\begin{equation}
\label{si}
 |\sigma( \bar y\cdot \eta+y_0)|\leq 1,\quad 
  |\sigma(\bar y\cdot \eta+y_0)-\sigma(\hat y\cdot \eta+\hat y_0)|\leq L(m\|\eta\|_{\ell_\infty^m}+1)\|y-y'\|_{\ell_\infty^{\tilde n}},
 \end{equation}
 and 
\begin{equation}
\label{sik}
  | (\bar y\cdot \eta+y_0)_+|\leq (m\|\eta\|_{\ell_\infty^m}+1)\|y\|_{\ell_\infty^{\tilde n}}\leq (m\|\eta\|_{\ell_\infty^m}+1)w(n).
\end{equation}
 In what follows, we use the notation
   $$
  \|g\|:=\max_{1\le i\le W} \|g_i\|_{C(\Omega)},
  $$
when working with   vector functions $g=(g_1,\dots,g_W)^T$ whose coordinates $g_i\in C(\Omega)$. 
  \begin{theorem}
  \label{NN}
 Let X be a Banach space such that $C([0,1]^d)\subset X$ is continuously embedded in $X$. Then the mapping $\Phi_\sigma:(B_{\ell_\infty^{\tilde n}}(w(n)), \|\cdot\|_{\ell_\infty^{\tilde n}})\to  X$, defined in 
  {\rm (\ref{NN1})}, is an $L_n$-Lipschitz mapping, that is
  $$
  \|\Phi_\sigma(y)-\Phi_\sigma(y')\|_X\leq c_0\|\Phi_\sigma(y)-\Phi_\sigma(y')\|_{C(\Omega)}\leq L_n\|y-y'\|_{\ell_\infty^{\tilde n}}, \quad  y,y'\in B_{\ell_\infty^{\tilde n}}(w(n)),
  $$ 
  where 
     the constant $L_n$ is bounded by
  $$
  2^{c_1n(1+\log_2w(n))}< L_n<2^{c_2n(1+\log_2w(n))},
  $$
  when $\sigma$ is an $L$-Lipschitz sigmoidal function or $\sigma=$ReLU and $LWw(n)\geq 2$. The constants  $c_1,c_2$ depend on  $c_0$, $d$, $L$, and $W$.
      \end{theorem}
  \noindent
 {\bf Proof:} The proof follows the arguments  from the proof of   Theorem 6.1 in \cite{PW}.  Let
    $y,y'$ be the two parameters from $B_{\ell_\infty^{\tilde n}}(w(n))$  that determine the 
    continuous functions $\Phi_\sigma(y), \Phi_\sigma(y')\in \Sigma_{\sigma,n}(w(n))$. They  are constructed by ordering in a predetermined way the entries of the affine mappings $A^{(j)}(\cdot):=A_j(\cdot)+b^{(j)}$,  $j=0,\ldots,n$,  and 
 $A'^{(j)}(\cdot):=A'_j(\cdot)+b'^{(j)}$,  $j=0,\ldots,n$, that define $\Phi_\sigma(y)$ and $\Phi_\sigma(y')$, respectively.
We fix $x\in \Omega$ and  denote  by 
 $$
  \eta^{(0)}(x) := { \overline\sigma} (A_0x+b^{(0)}),\quad  \eta'^{(0)}(x) := { \overline\sigma} (A'_0x+b'^{(0)}),
$$
$$  
\eta^{(j)}:= {\overline\sigma}(A_j\eta^{(j-1)}+b^{(j)}),\quad  \zeta'^{(j)}:= {\overline\sigma}(A'_j\eta'^{(j-1)}+b'^{(j)}), \quad j=1,\ldots,n-1,
$$
$$
\eta^{(n)} := A_n\eta^{(n-1)}+b^{(n)},\quad  \eta'^{(n)}:=A'_n\eta'^{(n-1)}+b'^{(n)},
$$
and 
 $$
  \zeta^{(0)}(x) := \overline{\rm ReLU}(A_0x+b^{(0)}),\quad  \zeta'^{(0)}(x) :=\overline{\rm ReLU}  (A'_0x+b'^{(0)}),
$$
$$  
\zeta^{(j)}:= \overline{\rm ReLU}(A_j\zeta^{(j-1)}+b^{(j)}),\quad  \zeta'^{(j)}:= \overline{\rm ReLU}(A'_j\zeta'^{(j-1)}+b'^{(j)}), \quad j=1,\ldots,n-1,
$$
$$
\zeta^{(n)} := A_n\zeta^{(n-1)}+b^{(n)},\quad  \zeta'^{(n)}:=A'_n\zeta'^{(n-1)}+b'^{(n)}.
$$
Note that  $A_0,A_0'\in \mathbb R^{W\times d}$, $A_j,A_j'\in \mathbb R^{W\times W}$, $b^{(0)},b'^{(0)},b^{(j)},b'^{(j)}\in \mathbb R^W$, for $j=1,\ldots,n-1$,  while
$A_n,A_n'\in \mathbb R^{1\times W}$, and $b^{(n)},b'^{(n)}\in \mathbb R$. Each of the $\eta^{(j)}, \eta'^{(j)}, \zeta^{(j)}, \zeta'^{(j)}$, $j=0, \ldots,n-1$, is a 
continuous vector function with $W$ coordinates and $\eta^{(n)}, \eta'^{(n)}, \zeta^{(n)}, \zeta'^{(n)}$ are  the outputs of the DNN with parameters $y,y'$ and activation function $\sigma$, and 
 $y,y'$ and activation function ReLU, respectively.
 
\underline {\bf Case 1:}  DNN with activation   function the Lipschitz  sigmoidal function $\sigma$.

\noindent
 Let us first observe, see  (\ref{si}),  that $\|\eta'^{(j)}\|\leq1$, $j=0, \ldots,n-1$, and 
 \begin{eqnarray}
\nonumber
 \|\eta^{(0)}-\eta^{'(0)}\|
 \leq L(d+1)\|y-y'\|_{\ell_\infty^{\tilde n}}
 =:C_0\|y-y'\|_{\ell_\infty^{\tilde n}}.
\end{eqnarray}
   Suppose we have proved that 
$$
 \|\eta^{(j-1)}-\eta'^{(j-1)}\|\leq C_{j-1}\|y-y'\|_{\ell_\infty^{\tilde n}}.
$$  
It follows that
  \begin{eqnarray} 
  \label{rec1}
  \nonumber
 \|\eta^{(j)}-\eta'^{(j)}\|&\leq& L \|A_j\eta^{(j-1)}+b^{(j)} -A_j'\eta'^{(j-1)}-b'^{(j)}\|
  \\ \nonumber
  &\le&   L\|A_j(\eta^{(j-1)}-\eta'^{(j-1)})\|+ L\|(A_j-A'_j)\eta'^{(j-1)}\|+L\|b^{(j)}-b'^{(j)}\|\\
  \nonumber
  &\leq &LW\|y\|_{\ell_\infty^{\tilde n}}\|\eta^{(j-1)}-\eta'^{(j-1)}\|
 +
LW\|y-y'\|_{\ell_\infty^{\tilde n}}\|\eta'^{(j-1)}\|+L\|y-y'\|_{\ell_\infty^{\tilde n}}\\
\nonumber
  &\le&
(LWw(n)C_{j-1}+LW+L)\|y-y'\|_{\ell_\infty^{\tilde n}}\\
\nonumber
& =:&
  C_{j}\|y-y'\|_{\ell_\infty^{\tilde n}},
  \end{eqnarray} 
  where  we  have used that $\|y\|_{\ell_\infty^{\tilde n}}\leq w(n)$,  $\|\eta'^{(j)}\|\leq 1$, and the induction hypothesis. 
 Thus, the relation between $C_j$ and $C_{j-1}$ is
   $$
  C_0=L(d+1), \quad  C_{j}=LWw(n)C_{j-1}+LW+L, \quad j=1,\ldots,n.
$$
  If we denote by $A:=LWw(n)$ and  $B:=LW+L=L(W+1)$, we have that
\begin{eqnarray}
\nonumber
 C_j&=&AC_{j-1}+B=\ldots=A^jC_0+(A^{j-1}+\ldots+1)B\\
 \nonumber
 &\leq &(A^j+\ldots+1)L(\max\{W,d\}+1)=\frac{A^{j+1}-1}{A-1}L(\max\{W,d\}+1)\\
 \nonumber
 &\leq &2A^jL(\max\{W,d\}+1)=C'[LWw(n)]^j, \quad C':=2L(\max\{W,d\}+1).
\end{eqnarray} 
  Finally, since
 $\|\Phi_\sigma(y)-\Phi_\sigma(y')\|_{C(\Omega)}=\|\eta^{(n)}-\eta'^{(n)}\|$, we have
 $$
 \|\Phi_\sigma(y)-\Phi_\sigma(y')\|_X\leq c_0\|\Phi_\sigma(y)-\Phi_\sigma(y')\|_{C(\Omega)}\leq c_0C_{n}\|y-y'\|_{\ell_\infty^{\tilde n}}<C'[LWw(n)]^{n}\|y-y'\|_{\ell_\infty^{\tilde n}}.
  $$
 The next case follows the same idea with several slight modifications.

 \underline{\bf Case 2:} DNN with activation   function ReLU.
 
 \noindent
   Note  that  since $\|y\|_{\ell_\infty^{\tilde n}}\leq w(n)$, it follows from (\ref{sik})  (for $m=d$ and $\eta=x$) that 
$$
\|\zeta'^{(0)}\|\leq dw(n)+w(n), 
$$
and (for $m=W$ and $\eta=\zeta'^{(j-1)}$)
$$
\|\zeta'^{(j)}\|\leq
   Ww(n)\|\zeta'^{(j-1)}\|+w(n), \quad j=1,\ldots,n.
  $$
 We want to point out that the last two inequalities hold even if we use  $\sigma(t)=t$ instead of $\sigma(t)={\rm ReLU}(t)$ for some of the coordinates in the definition \eref{bars} of $\bar \sigma$.

One can show by induction that for $j=1,\ldots,n$, 
\begin{eqnarray}
\nonumber
\|\zeta'^{(j)}\|&\leq &dW^jw(n)^{j+1}+w(n)\sum_{i=0}^j[Ww(n)]^i
\leq dW^jw(n)^{j+1}+2w(n)[Ww(n)]^j\\
\nonumber
&=&(d+2)w(n)[Ww(n)]^j,
\end{eqnarray}
since  $w(n)\geq 1\geq 2W^{-1}$.
Note that the above inequality also holds for $j=0$.
 Next, since  ReLU is a Lip 1 function,  see \eref{sik} with $L=1$, we have
\begin{eqnarray}
\nonumber
 \|\zeta^{(0)}-\zeta^{'(0)}\|
 \leq 
  (d+1)\|y-y'\|_{\ell_\infty^{\tilde n}}=:D_0\|y-y'\|_{\ell_\infty^{\tilde n}}.
\end{eqnarray}
   Suppose we have proved that 
$$
 \|\zeta^{(j-1)}-\zeta'^{(j-1)}\|\leq D_{j-1}\|y-y'\|_{\ell_\infty^{\tilde n}}.
$$  
Then,  similarly to Case 1,  we obtain that
  \begin{eqnarray} 
  \label{rec1}
  \nonumber
  \|\zeta^{(j)}-\zeta'^{(j)}\|&\le& 
 \|A_j(\zeta^{(j-1)}-\zeta'^{(j-1)})\|+ \|(A_j-A'_j)\zeta'^{(j-1)}\|+
\|b^{(j)}-b'^{(j)}\|\\
\nonumber
&\leq& W\|y\|_{\ell_\infty^{\tilde n}}\|\zeta^{(j-1)}-\zeta'^{(j-1)}\|+ W\|y-y'\|_{\ell_\infty^{\tilde n}}\|\zeta'^{(j-1)}\|+\|y-y'\|_{\ell_\infty^{\tilde n}}\\
\nonumber
  &\le&
(Ww(n)D_{j-1}+(d+2)w(n)[Ww(n)]^j+1)\|y-y'\|_{\ell_\infty^{\tilde n}}
  \\ \nonumber
  &=:&
  D_{j}\|y-y'\|_{\ell_\infty^{\tilde n}}.
  \end{eqnarray} 
   Thus,  $\|\zeta^{(j)}-\zeta'^{(j)}\|\leq D_j\|y-y'\|_{\ell_\infty^{\tilde n}}$, where 
\begin{eqnarray}
\nonumber
D_{j}&=&Ww(n)D_{j-1}+(d+2)w(n)[Ww(n)]^j+1.
\end{eqnarray}
Since $D_0=d+1$ and 
$$
D_1=Ww(n)D_0+(d+2)Ww(n)^2+1<(d+2)(Ww(n)^2+Ww(n)+1),
$$
 we obtain by induction that 
\begin{eqnarray}
\nonumber
  D_{n}&<&(d+2)\left(nw(n)[Ww(n)]^{n}+\sum_{i=0}^{n}[Ww(n)]^{i}\right)
 < (d+2)(nw(n)+2)[Ww(n)]^{n}
  \\
  \nonumber
  &<& (d+2)(nw(n)+Ww(n))[Ww(n)]^{n}<
W(d+2)nw(n)[Ww(n)]^{n},
\end{eqnarray}
 since  $L=1$ in this case and $Ww(n)\geq 2$. 
Finally, we have
 $$
\|\Phi_\sigma(y)-\Phi_\sigma(y')\|_{C(\Omega)}=\|\zeta^{(n)}-\zeta'^{(n)}\| \leq D_{n}\|y-y'\|_{\ell_\infty^{\tilde n}}<(d+2)n[Ww(n)]^{n+1}\|y-y'\|_{\ell_\infty^{\tilde n}}.
  $$
 In both Case 1 and Case 2, the Lipschitz constant $L_n$ is such that we can find constants $c_1,c_2>0$ with the property
 $$
 2^{c_1n(1+\log_2w(n)})<L_n<2^{c_2n(1+\log_2w(n))}
 $$
 and the proof is completed.
\hfill $\Box$

\bigskip
\begin{remark}
\label{Sremark}
Note that one can follow  the  same arguments as in Case {\rm 2} and prove Theorem {\rm \ref{NN}} when every coordinate  of $\bar \sigma$, see 
\eref{bars}, is chosen to be either  
 $\sigma(t)={\rm ReLU}(t)$ or $\sigma(t)=t$,  and this choice can change from layer to layer.
 \end{remark}

 \subsection{Shallow neural networks}
 In this section we consider SNNs and prove that they are also Lipschitz mappings. The following theorem holds.
\begin{theorem}
  \label{NNs}
 Let X be a Banach space such that $C([0,1]^d)\subset X$ is continuously embedded in $X$. Then the  mapping $\Psi_\sigma:(B_{\ell_\infty^{\widetilde W}}(w(W)), \|\cdot\|_{\ell_\infty^{\widetilde W}})\to X$, defined in 
  {\rm (\ref{NNsdef})}, is an $L_W$-Lipschitz mapping, that is
  $$
  \|\Psi_\sigma(y)-\Psi_\sigma(y')\|_X\leq c_0 \|\Psi_\sigma(y)-\Psi_\sigma(y')\|_{C(\Omega)}\leq L_W\|y-y'\|_{\ell_\infty^{\widetilde W}}, \quad  y,y'\in B_{\ell_\infty^{\widetilde W}}(w(W)),
  $$ 
  with  constant 
  $$L_W=\begin{cases}
  CWw(W), \quad \sigma \,\,\hbox{is L-Lipschitz sigmoidal function},\\
  CW[w(W)]^2, \quad\sigma=\hbox{ReLU},
  \end{cases}
  $$
 where $C=C(d,L,c_0)$ and  $w(W)\geq 1$.
    \end{theorem}
    \noindent
{\bf Proof:} We follow the proof of  Theorem \ref{NN}. In  Case 1 we have 
$$
L_W=c_0C_1=c_0(LWw(W)(d+1)+LW+L)\leq CWw(W),\quad C= c_0(L(d+1)+2L),
$$
 and 
in Case 2 we have 
$$
L_W=c_0D_1<c_0(d+2)(Ww(W)^2+Ww(W)+1)\leq CWw(W)^2,\quad C=3c_0(d+2),
$$
provided   $w(W)\geq 1$.
\hfill $\Box$


\section{Lipschitz widths}
\label{S3}
The Lipschitz widths  $d_n^{\gamma}({\mathcal K})_X$ of a compact subset $\mathcal K\subset X$  of a Banach space $X$ were introduced  in \cite{PW}  
via Lipschitz mappings defined on unit balls $B_{Y_n}(1)$. 
However, having in mind  approximation via NNs with  parameters that are bounded by $w(n)$, we need to consider Lipschitz mappings whose domain are balls
$B_{Y_n}(w(n))$ with radius $w(n)$. The next lemma shows how the Lipschitz width $d_n^{\gamma}({\mathcal K})_X$ is related to  all $\gamma/r$-Lipschitz mappings 
${\mathcal L}_n$ with domain  $B_{Y_n}(r)$, $r>0$,  whose  image is  in   $X$.
 More precisely, we prove that in the definition of fixed Lipschitz widths we can consider mappings defined on balls with changing radiuses as long as the product of the 
 Lipschitz constant of the mappings and the radius of the ball does not exceed $\gamma$.
\begin{lemma}
\label{Lemma1}
For any compact subset ${\mathcal K}$ of $X$, any $\gamma>0$, any $n\geq 1$, and any norm $\|\cdot\|_{Y_n}$ on $\mathbb R^n$, we have that 
\begin{equation}
\label{pp1}
d^\gamma({\mathcal K},Y_n)_X:=\inf_{{\mathcal L}_n, r>0}\,\,\sup_{f\in {\mathcal K}}\,\,\inf_{y\in B_{Y_n}(r)} \|f-{\mathcal L}_n(y)\|_X,
\end{equation}
where the  infimum is taken over all $\gamma/r$-Lipschitz mappings 
${\mathcal L}_n:(B_{Y_n}(r),\|\cdot\|_{Y_n})\to X$,  and all $r>0$.
 \end{lemma}
\noindent
{\bf Proof:}
We fix a  norm $\|\cdot\|_{Y_n}$ on $\mathbb R^n$  and take  any $r>0$ and  map  ${\mathcal L}_n:(B_{Y_n}(r),\|\cdot\|_{Y_n})\to X$
with  Lipschitz constant $\gamma/r$.
We then consider the onto $r$-Lipschitz  map $\eta:B_{Y_n}(1) \rightarrow B_{Y_n}(r)$, defined as $\eta(y)=ry$.
Clearly,   ${\mathcal L}_n\circ \eta:(B_{Y_n}(1),\|\cdot\|_{Y_n})\to X$ is  $\gamma$-Lipschitz.
Therefore 
we have
$$
d^{\gamma}({\mathcal K}, Y_n)_X\leq  \sup_{f\in{\mathcal K}}\,\,\inf_{y\in B_{Y_n}(1)} \|f-{\mathcal L}_n\circ \eta(y)\|_X=\sup_{f\in{\mathcal K}}\,\,\inf_{y\in B_{Y_n}(r)} \|f-{\mathcal L}_n(y)\|_X,
$$
 which gives
\begin{equation}
\label{p2}
d^{\gamma}({\mathcal K}, Y_n)_X\leq \inf_{{\mathcal L}_n,  r>0}\,\,\sup_{f\in{\mathcal K}}\,\,\inf_{y\in B_{Y_n}(r)} \|f-{\mathcal L}_n(y)\|_X,
\end{equation}
where the infimum  is  taken over    all  $r>0$ and all $\gamma/r$-Lipschitz maps  ${\mathcal L}_n:(B_{Y_n}(r),\|\cdot\|_{Y_n})\to X$.
 
Observe that we can argue in the another direction too. For fixed $\delta>0$, we take  a $\gamma$-Lipschitz mapping ${\mathcal L}'_n:(B_{Y_n}(1),\|\cdot\|_{Y_n})\to X$ such that 
$$
\sup_{f\in {\mathcal K}}\,\,\inf_{y\in B_{Y_n}(1)}\|f-{\mathcal L}'_n(y)\|_X \leq d^\gamma({\mathcal K},Y_n)_X+\delta.
$$
 We fix $r>0$, define the $1/r$-Lipschitz mapping $\eta$ from $B_{Y_n}(r)$ onto $B_{Y_n}(1)$ as $\eta(y)=y/r$ and
 consider the $\gamma/r$-Lipschitz mapping ${\mathcal L}'_n\circ \eta:(B_{Y_n}(r),\|\cdot\|_{Y_n})\mathbb \rightarrow X$.
Since  ${\mathcal L}'_n\circ \eta$ is a $\gamma/r$-Lipschitz mapping  and we have
 $$
 \inf_{{\mathcal L}_n,r>0}\,\,\sup_{f\in {\mathcal K}\,\,\inf_{y\in B_{Y_n}(r)} \|f-{\mathcal L}_n(y)\|_X\leq}  \sup_{f\in {\mathcal K}}\,\,\inf_{y\in B_{Y_n}(r)}\| f-{\mathcal L}'_n\circ \eta(y)\| = \sup_{f\in {\mathcal K}}\,\,\inf_{y\in B_{Y_n}(1)}\| f-
{\mathcal L'_n}(y)\|,
  $$
  which gives that
  $$
\inf_{{\mathcal L}_n,r>0}\sup_{f\in {\mathcal K}}\,\,\inf_{y\in B_{Y_n}(r)} \|f-{\mathcal L}_n(y)\|_X \leq d^\gamma({\mathcal K},Y_n)_X+\delta.
  $$
Since $\delta$ is arbitrary,  the above inequality and (\ref{p2}) prove the lemma.
\hfill $\Box$


\section{ Bounds for Lipschitz widths and  the limitations of NN approximation}
\label{LB}
 In this section we obtain  lower bounds for Lipschitz widths with large Lipschitz constants and apply them in the case of deep and shallow NNs.
\subsection{Lower bounds for Lipschitz widths}
\label{LB1}
We start our investigation with the observation, that the Lipschitz width $d_n^{\gamma}(\mathcal K)_X$,  $\gamma>0$, is bounded from below by the fixed width 
$d_n^{(2n+1)\gamma}(\mathcal K,{Z})_X$ with respect to any chosen in advance norm $\|\cdot\|_Z$ on $\mathbb R^n$ by paying the price of the slightly bigger 
Lipschitz constant $(2n+1)\gamma$. Note that  $d_n^{\gamma}(\mathcal K)_X$ is defined as infimum over all norms in $\mathbb R^n$, and thus the lemma below provides 
a way for calculating  lower estimates for the Lipschitz width since we can perform our computations using our favorable (easy to handle) norm  $Z$.
\begin{lemma}
\label{L2a}
For any compact subset ${\mathcal K}$ of $X$, any $\gamma>0$, any $n\geq 1$, and any  { norm $\|\cdot\|_Z$ on $\mathbb R^n$} we have that 
$$
d_n^{(2n+1)\gamma}(\mathcal K,{Z})_X\leq d_n^{\gamma}(\mathcal K)_X.
$$
\end{lemma}
\noindent
{\bf Proof:}  It is known   (see  \cite[Prop. 37.1]{TJ})  that for any two  $n$-dimensional Banach spaces $Y_1$ and $Y_2$ there exists a constant $\rho\leq n$,
$\rho=\rho(n)$, and  an onto  linear map $T:Y_1\rightarrow Y_2$ such that $\|T\|\cdot\|T^{-1}\|=\rho$.  By suitable rescaling we can assume that 
{  $\|T\|=1$ and $\|T^{-1}\|=\rho$. 

Let $(\mathbb R^n,\|.\|_{\mathcal Y_n})$ be the  space determined from the norm $\|\cdot\|_{{\mathcal Y}_n}$ in Theorem \ref{specialnorm}. From the above we infer that there exists a linear map $T$ with the properties
$$
T:(\mathbb R^n,\|.\|_Z) \rightarrow (\mathbb R^n,\|.\|_{\mathcal Y_n}), \quad \|T\|=1, \quad \|T^{-1}\|=\rho.
$$
We now define the mapping 
$$
\phi(y):=t(y)Ty, \quad \hbox{where}\quad t(y):=\frac{\|y\|_Z}{\|Ty\|_{ \mathcal Y_n}}.
$$
Since $\|\phi(y)\|_{ \mathcal Y_n}=\|y\|_{Z}$, we conclude that 
 $\phi:(B_{Z}(1),\|\cdot\|_{Z})\to(B_{ \mathcal Y_n}(1),\|\cdot\|_{ \mathcal Y_n})$. Moreover,  $\phi$ is  an onto mapping since
for every $y'\in B_{ \mathcal Y_n}(1)$, there is 
$$
y=\frac{\|y'\|_{\mathcal Y_n}}{\|T^{-1}y'\|_Z}T^{-1}y'\in B_{Z}(1),\quad \hbox{ such that}\quad  \phi(y)=y'.
$$
Note that 
$t(y)\leq \rho$ since
$
\|y\|_Z=\|T^{-1}(Ty)\|_Z\leq \|T^{-1}\|\|Ty\|_{\mathcal Y_n}=\rho\|Ty\|_{\mathcal Y_n},
$
and
\begin{eqnarray*}
\nonumber
|t(y)-t(z)|&\leq& \left |  \frac{\|y\|_{Z}}{\|Ty\|_{\mathcal Y_n}} -\frac{\|y\|_{Z}}{\|Tz\|_{\mathcal Y_n}} \right |+\left |\frac{\|y\|_{Z}}{\|Tz\|_{\mathcal Y_n}}- \frac{\|z\|_{Z}}{\|Tz\|_{\mathcal Y_n}}\right |\\
\nonumber
&=&\frac{\|y\|_{Z}}{\|Ty\|_{\mathcal Y_n}\|Tz\|_{\mathcal Y_n}}|\|Tz\|_{ \mathcal Y_n}-\|Ty\|_{\mathcal Y_n}|
+\frac{1}{\|Tz\|_ {\mathcal Y_n}}\left|\|z\|_{Z}-\|y\|_{Z}\right|\\
\label{w1}
& \leq &\frac{t(y)}{\|Tz\|_{\mathcal Y_n}}\|T(z-y)\|_{ \mathcal Y_n}+\frac{1}{\|Tz\|_{\mathcal Y_n}}\|z-y\|_{Z}
\leq  \frac{{\rho}+1}{\|Tz\|_{ \mathcal Y_n}}\|z-y\|_{Z}.
\end{eqnarray*}
Therefore, for every $y,z\in B_{Z}(1)$, using the above two inequalities, we obtain that 
\begin{eqnarray*}
\|\phi(y)-\phi(z)\|_{ \mathcal Y_n}&\leq&t(y)\|T(y-z)\|_{\mathcal Y_n}+|t(y)-t(z)|\|Tz\|_{\mathcal Y_n} \\
&\leq& \rho\|y-z\|_{Z}+(\rho+1)\|y-z\|_{Z}
=(2\rho+1)\|y-z\|_{Z},
\end{eqnarray*}
 which shows that  $\phi$ is a $(2\rho+1)$-Lipschitz  mapping.

Now, if ${\mathcal L}_n:(B_{ \mathcal Y_n}(1),\|\cdot\|_{ \mathcal Y_n})\to X$ is any $\gamma$-Lipschitz mapping, then 
${\mathcal L}_n\circ \phi:(B_{Z}(1),\|\cdot\|_Z)\to X$
is a $(2\rho+1)\gamma$-Lipschitz mapping for which 
$$
d_n^{(2\rho+1)\gamma}(\mathcal K,Z)_X\leq \sup_{f\in{\mathcal K}}\,\,\inf_{y\in B_{Z}(1)}\|f-{\mathcal L}_n\circ \phi(y)\|_X=\sup_{f\in{\mathcal K}}\,\,\inf_{y\in B_{ \mathcal Y_n}(1)}\|f-{\mathcal L}_n(y)\|_X.
$$
Next, we take  the infimum  over $\Phi_n$ and obtain 
 $$
    d_n^{(2\rho+1)\gamma}(\mathcal K,Z)_X  \leq   d_n^\gamma(\mathcal K,\mathcal Y_n)_X =d_n^\gamma(\mathcal K)_X,
    $$
 which completes the proof since $\rho\leq n$ and the Lipschitz width is monotone with respect to the Lipschitz constant.
    \hfill $\Box$
    
 In the above inequality we may choose the $Z$ norm to be the $\ell_2^n$ norm, in  which  case $\rho\leq \sqrt{n}$. Thus, we obtain the following remark.
\begin{remark}
\label{R1}
For any compact subset ${\mathcal K}$ of $X$, any $\gamma>0$, and any $n\geq 1$, we have that 
$$
d_n^{(2\sqrt{n}+1)\gamma}(\mathcal K)_X\leq d_n^{(2\sqrt{n}+1)\gamma}(\mathcal K,\ell_2^n)_X\leq d_n^{\gamma}(\mathcal K)_X.
$$
\end{remark}

\bigskip

We now proceed with the investigation of other lower bounds for the Lipschitz widths by first  recalling the following proposition, see   \cite[Prop 3.8]{PW}.
\begin{prop}\label{carl2} 
Let  $\mathcal K\subset X$ be a compact set and let 
$\epsilon_n(\mathcal K)_X> \eta_n$, $n=1,2,\dots$,
 where  $(\eta_n)_{n=1}^\infty$ is a sequences of real numbers decreasing to zero.  
If  for some $m\in \mathbb N$ { and  $\delta>0$} we have that 
$ d_{m}^\gamma(\mathcal K)_X<\delta$,
then 
\begin{equation}
\label{carl3}
{ \eta_{m\log_2 (3\gamma\delta^{-1})} { <}2 \delta.}
\end{equation}
\end{prop}
This proposition was used  in  \cite{PW}   to prove lower bounds for the Lipschitz widths $d_n^{\gamma_n}({\mathcal K})_X$  of the compact set $\mathcal K$
in the cases $\gamma_n=2^{\varphi(n)}$, 
  $\varphi(n)=const$ (see \cite[Theorem  3.9]{PW}) and  $\varphi(n)=c'n$ (see \cite[Theorem  6.3]{PW}),  provided we have information 
 about the entropy numbers of  $\mathcal K$. The next 
theorem  is a generalization of Theorem  6.3 from \cite{PW} for the case of a general function $\varphi$.

\begin{theorem}
\label{widthsfrombelownew}  
For any  compact set ${\mathcal K}\subset X$, we consider the Lipschitz width $d_n^{\gamma_n}({\mathcal K})_X$ with 
 Lipschitz constant  $\gamma_n=2^{\varphi(n)}$, where   $\varphi(n)\geq c\log_2n$ for some fixed constant $c>0$. Then the following holds:
\begin{equation} 
\label{widths(i)gen}
{\rm (i)}\quad \epsilon_n({\mathcal K})_X\gtrsim \frac{(\log_2 n)^\beta}{n^{\alpha}},\quad n\in \mathbb N\quad  \Rightarrow\quad d_n^{\gamma_n}({\mathcal K})_X\gtrsim[\log_2 (n\varphi(n))]^{\beta}[n\varphi(n)]^{-\alpha}, \quad n\in \mathbb N,
\end{equation}
 where $\alpha>0$, $\beta\in \mathbb R$.
\begin{equation} 
\label{widths(ii)gen}
{\rm (ii)}\quad\quad\quad\quad\quad\quad\epsilon_n({\mathcal K}) _X\gtrsim [\log_2 n]^{-\alpha},\quad n\in \mathbb N\Rightarrow\quad d_n^{\gamma_n}({\mathcal K})_X\gtrsim [\log_2 (n\varphi(n))]^{-\alpha}, \quad n\in \mathbb N,
\end{equation} 
where $\alpha>0$.
\end{theorem}
\noindent 
{\bf Proof:} We provide the proof in the Appendix.
\hfill $\Box$
\bigskip

\subsection{Lower bounds  for DNN approximation}

 In this section we consider Banach spaces $X$ such that  $C([0,1]^d)$ in continuously embedded in $X$. 
Let us denote by 
$$
E(f,\Sigma_{0,\sigma}(w(n)))_X:=\|f\|_X,\quad E(f,\Sigma_{n,\sigma}(w(n)))_X:=\inf_{y\in B_{\ell_\infty}^{\tilde n}(w(n))}\|f-\Phi_\sigma(y)\|_X,\quad n\geq 1,
$$
the error of approximation of a function $f$ by the outputs  $\Phi_\sigma(y)\in \Sigma_{\sigma,n}(w(n))$ of the DNN with parameters $y$, for which $\|y\|_{\ell_\infty^{\tilde n}}\leq w(n)$, $LWw(n)\geq 2$, measured
  in $\|\cdot\|_X$. We then denote by 
$$
E(\mathcal K,\Sigma_{n,\sigma}(w(n)))_X:=\sup_{f\in \mathcal K}\,\,E(f,\Sigma_{n,\sigma}(w(n)))_X,\quad n\geq 0,
$$
the error for the class $\mathcal K$.
 It follows from Lemma \ref{Lemma1} and Theorem \ref{NN}  that 
\begin{equation}
\label{main}
E(\mathcal K,\Sigma_{n,\sigma}(w(n)))_X\geq d^{\gamma_n}_n(\mathcal K)_X, \quad \hbox{with}\quad \gamma_n=2^{cn(1+\log_2w(n))}=:2^{\varphi(n)}, \quad c>0.
\end{equation}
The latter estimate shows that lower bounds for the error $E(\mathcal K,\Sigma_{n,\sigma}(w(n)))_X$ can be obtained by using  lower bounds for $d^{\gamma_n}_n(\mathcal K)_X$,
 and thus provides a way to study the   theoretical  limitations of DNNs with ReLU and Lipschitz sigmoidal activation functions and $w(n)$ bounds for their parameters.  
 We next apply the results obtained in \S\ref{LB1} to the special  case of DNNs.
\begin{remark}
\label{RNN}
It follows from \eref{main}, Lemma {\rm \ref {L2a}} with $\gamma_n=2^{cn(1+\log_2w(n))}$, and the monotonicity with respect to $\gamma$ of the fixed Lipschitz width   
that  for any compact subset ${\mathcal K}$ of a Banach space $X$,    
$$
E(\mathcal K,\Sigma_{n,\sigma}(w(n)))_X\geq d_n^{\gamma_n}(\mathcal K)_X\geq d_n^{(2n+1)\gamma_n}(\mathcal K,{Z})_X\geq d_n^{2^{c_1n(1+\log_2w(n))}}(\mathcal K,{Z})_X,
$$
where  $\|\cdot\|_Z$ is  any norm on $\mathbb R^n$.
\end{remark}
The following Table 1 shows the relation (for sufficiently large $n$) between the bound $w(n)$ and the 
parameter  $\gamma_n=2^{\varphi(n)}$
of the Lipschitz width $d^{\gamma_n}_n(\mathcal K)_X$ from \eref{main}, where $\varphi(n)=cn(1+\log_2w(n))$. 
\begin{table}[h]
\label{T1}
    \centering 
\begin{tabular}{||c| c|c||} 
 \hline
 $w(n)$ & $\varphi(n)=cn(1+\log_2w(n))$&$\gamma_n=2^{\varphi(n)}$\\ [0.5ex] 
 \hline\hline
$C\geq 1$ &$c'n$, $\quad c'>0$&$\lambda^n$, $\quad\lambda>1$\\ 
 \hline
$Cn^\delta$, $\quad C,\delta>0$& $c'n\log_2n$, $\quad c'>0$&$2^{c'n\log_2n}$, $\quad c'>0$\\
 \hline
 $C2^{cn^\nu}$, $\quad C,c,\nu>0$&$c'n^{\nu+1}$, $\quad c'>0$& $2^{c'n^{\nu+1}}$, $\quad c'>0$ \\
 \hline
\end{tabular}
    \caption{Relation between $w(n)$, $\varphi(n)$ and $\gamma_n$.}
\end{table}

\newpage
The next corollary follows from \eref{main} and Theorem \ref{widthsfrombelownew} when $\varphi(n)=cn(1+\log_2w(n))$, $c>0$.

\begin{cor}
\label{mainc}
The error of approximation  of a compact subset ${\mathcal K}$ of a Banach space $X$ by NNs with depth $n$, Lipschitz sigmoidal or ReLU activation function and $w(n)$ bound on its parameters,
with $LWw(n)\geq 2$, 
satisfies the following lower bounds,  provided we have certain lower bounds for the entropy numbers:
\begin{equation} 
\nonumber
\epsilon_n({\mathcal K})_X\gtrsim  \frac{(\log_2 n)^\beta}{n^{\alpha}}, \, n\in \mathbb N\quad \Rightarrow \quad E(\cK,\Sigma_{n,\sigma}(w(n)))_X\gtrsim \frac{[\log_2 n+\log_2(1+\log_2w(n))]^{\beta}}{[n^2(1+\log_2w(n))]^{\alpha}},
 \, n\in \mathbb N,
\end{equation} 
\begin{equation} 
\nonumber
\epsilon_n({\mathcal K}) _X\gtrsim [\log_2 n]^{-\alpha}, \,n\in \mathbb N\quad \Rightarrow\quad E(\cK,\Sigma_{n,\sigma}(w(n)))_X\gtrsim [\log_2 n+\log_2(1+\log_2w(n))]^{-\alpha}, \, n\in \mathbb N.
\end{equation} 
In particular, if $w(n)=Cn^\delta$,  with $\delta>0$, we have:   
$$
 \epsilon_n({\mathcal K})_X\gtrsim [\log_2 n]^\beta n^{-\alpha}\quad \Rightarrow \quad E(\mathcal K,\Sigma_{n,\sigma}(Cn^\delta))_X\gtrsim[\log_2 n]^{\beta-\alpha}n^{-2\alpha},
$$
$$
\epsilon_n({\mathcal K}) _X\gtrsim[\log_2 n]^{-\alpha}\quad\Rightarrow \quad E(\mathcal K,\Sigma_{n,\sigma}(Cn^\delta))_X\gtrsim[\log_2 n]^{-\alpha},
$$
 and when $w(n)=C2^{cn^\nu}$,  with $C,c>0$, $\nu\geq 0$, we have: 
$$
 \epsilon_n({\mathcal K})_X\gtrsim [\log_2 n]^\beta n^{-\alpha}\quad \Rightarrow \quad E(\mathcal K,\Sigma_{n,\sigma}(C2^{cn^\nu})_X\gtrsim[\log_2 n]^{\beta}n^{-(2+\nu)\alpha},
 $$
 $$
\epsilon_n({\mathcal K}) _X\gtrsim[\log_2 n]^{-\alpha}\quad \Rightarrow \quad E(\mathcal K,\Sigma_{n,\sigma}(C2^{cn^\nu})_X\gtrsim[\log_2 n]^{-\alpha}.
$$
\end{cor}

\subsection{Lower bounds for  shallow neural network approximation}
\label{SH}

 In this section we consider Banach spaces $X$ such that  $C([0,1]^d)$ in continuously embedded in $X$. 
Let us denote by
$$
E(f, \Xi_{j,\sigma}(w(W)))_X:=\|f\|_X,\quad j=0,1,
$$
$$
 E(f,\Xi_{n,\sigma}(w(W)))_X:=\inf_{y\in B_{\ell_\infty}^{\tilde W}(w(W))}\|f-\Psi_\sigma(y)\|_X,\quad W\geq 2,
$$
the error of approximation of a function $f$ by the outputs  $\Psi_\sigma(y)\in \Xi_{W,\sigma}(w(W))$ of the DNN with parameters $y$, for which $\|y\|_{\ell_\infty^{\widetilde W}}\leq w(W)$, 
$w(W)\geq 1$,
measured  in the norm of the Banach space $X$,
and by 
$$
E(\mathcal K, \Xi_{W,\sigma}(w(W)))_X:=\sup_{f\in \mathcal K}\,\,E(f, \Xi_{W,\sigma}(w(W)))_X,\quad W\geq 0,
$$
the error for the class $\mathcal K$.
Note that in the case of SNNs, the sets $\Xi_{W,\sigma}(w(W))$ are nested, namely 
 $$
 \Xi_{W,\sigma}(w(W))\subset \Xi_{W+1,\sigma}(w(W)),
 $$
 and therefore
 $$
 E(f,\Xi_{W+1,\sigma}(w(W)))_X\leq E(f,\Xi_{W,\sigma}(w(W)))_X.
 $$

 It follows from Lemma \ref{Lemma1} and Theorem \ref{NNs}  that 
\begin{equation}
\label{mains}
E(\mathcal K,\Xi_{W,\sigma}(w(W)))_X\geq d^{\gamma_W}_W(\mathcal K)_X, \quad \hbox{with}\quad 
\gamma_W=
2^{c(\log_2W+\log_2w(W))}=:2^{\varphi(n)},
\end{equation}
and thus  lower bounds for the error $E(\mathcal K,\Xi_{W,\sigma}(w(W)))_X$ can be obtained by using  lower bounds for $d^{\gamma_W}_W(\mathcal K)_X$.
We next apply the results obtained in \S\ref{LB1} to the special  case of SNNs.
\begin{remark}
\label{RNNs}
It follows from \eref{mains}, Lemma {\rm \ref {L2a}} with $\gamma_W=2^{c(\log_2W+\log_2w(W))}$, and the monotonicity with respect to $\gamma$ of the fixed Lipschitz width   
that  for any compact subset ${\mathcal K}$ of a Banach space $X$,   
$$
E(\mathcal K,\Xi_{W,\sigma}(w(W)))_X\geq d_W^{\gamma_W}(\mathcal K)_X\geq d_W^{(2W+1)\gamma_W}(\mathcal K,{Z})_X\geq d_W^{2^{c_1W(\log_2W+\log_2w(W))}}(\mathcal K,{Z})_X, \quad c_1>0,
$$
where  $\|\cdot\|_Z$ is  any norm on $\mathbb R^W$.
\end{remark}

As in the case of DNNs, we create a table showing the relation (for sufficiently large $W$) between the bound $w(W)$ and the 
parameter  $\gamma_W=2^{\varphi(W)}$
of the Lipschitz width $d^{\gamma_W}_W(\mathcal K)_X$ from \eref{mains}, where $\varphi(W)=c(\log_2W+\log_2w(W))$. 
\begin{table}[h]
\label{T2}
    \centering 
\begin{tabular}{||c| c|c||} 
 \hline
 $w(W)$ & $\varphi(W)=c(\log_2W+\log_2w(W))$&$\gamma_W=2^{\varphi(W)}$\\ [0.5ex] 
 \hline\hline
$CW^\delta$, $\quad C,\delta\geq 0$ &$c'\log_2W$, $\quad c'>0$&$2^{c'\log_2W}$\\ 
 \hline
 $C2^{cW^\nu}$, $\quad C,c,\nu>0$&$c'W^{\nu}$, $\quad c'>0$& $2^{c'W^{\nu}}$, $\quad c'>0$ \\
 \hline
\end{tabular}
    \caption{Relation between $w(W)$, $\varphi(W)$ and $\gamma_W$, shallow NNs}
\end{table}

\noindent
The next corollary follows from \eref{mains} and Theorem \ref{widthsfrombelownew} when $\varphi(W)=c(\log_2W+\log_2w(W))$, $c>0$.

\begin{cor}
\label{maincs}
The error of approximation  of a compact subset ${\mathcal K}$ of a Banach space $X$ by SNNs with width  $W$, Lipschitz sigmoidal or ReLU activation function and $w(W)\geq 1$ bound on its parameters
satisfies the following lower bounds,  provided we have certain lower bounds for the entropy numbers:
\begin{equation} 
\nonumber
\epsilon_n({\mathcal K})_X\gtrsim  \frac{(\log_2 n)^\beta}{n^{\alpha}}, \, n\in \mathbb N\quad \Rightarrow \quad 
E(\cK,\Xi_{W,\sigma}(w(W)))_X\gtrsim \frac{[\log_2 W+\log_2(\log_2W+\log_2w(W))]^{\beta}}{[W(\log_2W+\log_2w(W))]^{\alpha}}, 
\end{equation}
\begin{equation} 
\nonumber
\epsilon_n({\mathcal K}) _X\gtrsim [\log_2 n]^{-\alpha}, \,n\in \mathbb N\quad \Rightarrow\quad 
E(\cK,\Xi_{W,\sigma}(w(W)))_X\gtrsim [W(\log_2W+\log_2w(W))]^{-\alpha}, \, W\in \mathbb N.
\end{equation} 
In particular, if $w(W)=CW^\delta$,  with $\delta\geq 0$, we have:   
$$
 \epsilon_n({\mathcal K})_X\gtrsim [\log_2 n]^\beta n^{-\alpha}, n\in \mathbb N\quad \Rightarrow \quad E(\mathcal K,\Xi_{W,\sigma}(CW^\delta))_X\gtrsim[\log_2 W]^{\beta-\alpha}W^{-\alpha},  \, W\in \mathbb N,
 $$
 $$
\epsilon_n({\mathcal K}) _X\gtrsim[\log_2 n]^{-\alpha}, n\in \mathbb N\quad \Rightarrow \quad E(\mathcal K,\Xi_{W,\sigma}(CW^\delta))_X\gtrsim[\log_2 W]^{-\alpha}, \, W\in \mathbb N,
$$
 and when $w(W)=C2^{cW^\nu}$,  with $C\geq 1,\,c>0$, $\nu > 0$, we have: 
$$
 \epsilon_n({\mathcal K})_X\gtrsim [\log_2 n]^\beta n^{-\alpha}, n\in \mathbb N\quad \Rightarrow \quad E(\mathcal K,\Xi_{W,\sigma}( C2^{cW^\nu})_X\gtrsim[\log_2 W]^{\beta}W^{-(1+\nu)\alpha}, \, W\in \mathbb N,
 $$
$$
\epsilon_n({\mathcal K}) _X\gtrsim[\log_2 n]^{-\alpha}, n\in \mathbb N\quad \Rightarrow \quad E(\mathcal K,\Xi_{W,\sigma}( C2^{cW^\nu})_X\gtrsim[\log_2 W]^{-\alpha}, \, W\in \mathbb N.
$$
\end{cor}

 \begin{remark}
A version of  Corollary {\rm \ref{maincs}} was proven in Corollary  {\rm 2} from {\rm \cite {JS}}  in the case of SNN with ${\rm ReLU}^k$  activation function, $X=L_2(\Omega)$,
  $\xi_W=CW^{-\alpha}$, and $w(W)=C$, for classes 
${\mathcal K}=\overline{\{f=\sum_{j=1}^nc_jh_j,\,\,h_j\in\mathcal P_k^d, \,\,\sum_{j=1}^n|a_j|\leq 1\}}$, which is the closure of the convex, symmetric hull of
$\mathcal P_k^d:=\{{\rm ReLU}^k(\omega\cdot x+b):\,\omega\in S^{d-1}, \,b\in [c_1,c_2]\subset \mathbb R\}$.
\end{remark}


\subsection{Further study of  Lipschitz widths with large Lipschitz constants, part I}
\label{p1}
So far, we have used Lipschitz widths as a tool to obtain lower bounds for the error of approximation of a compact set $\cK$ via deep and shallow  NNs,  but Lipschitz widths are a subject of  interest on their own.
We have studied  in \cite{PW} Lipschitz widths with  Lipschitz constants $\gamma=const$ and $\gamma=\gamma_n=\lambda^n$, $\lambda>1$.  In this section, we will complete this study by including 
Lipschitz widths with constants $\gamma_n=2^{\varphi(n)}$. We start with the following theorem, which is an application  of Theorem 3.3 from \cite{PW}.

\begin{theorem}
\label{cr1}
Let  ${\mathcal K}\subset X$ be a  compact subset of a Banach space $X$,  $n\in \mathbb N$, and $d_n^{\gamma_n}({\mathcal K})_X$ be the   Lipschitz width for ${\mathcal K}$ with Lipschitz constant   $\gamma_n=2^{\varphi(n)}$, where $\varphi(n)\to \infty$ as $n\to\infty$. 
Then,  we have
\begin{equation}
\label{gg}
d^{2^{\varphi(n)}}_{n}({\mathcal K})_X\leq  \epsilon_{ n\lceil\frac{\varphi(n)}{2}\rceil}({\mathcal K})_X, \quad \hbox{where}\quad n\geq n_0(\mathcal K).
\end{equation}
In particular, 
\begin{eqnarray}
\label{pq}
{\rm (i)}\quad \epsilon_n({\mathcal K})_X\lesssim[\log_2n]^\beta n^{-\alpha}, \,n\in \mathbb N\quad \Rightarrow\quad d_n^{\gamma_n}({\mathcal K})_X\lesssim[\log_2(n\varphi(n))]^{\beta}[n\varphi(n)]^{-\alpha}, \,n\in\mathbb N,
\end{eqnarray}
where  $\alpha>0,\,\beta\in\mathbb R$,
\begin{equation}
\label{pq1}
{\rm (ii)}\quad\quad\quad \epsilon_n({\mathcal K})_X\lesssim [\log_2 n]^{-\alpha}, \,n\in \mathbb N,\,\alpha>0\quad \Rightarrow\quad d_n^{\gamma_n}({\mathcal K})_X\lesssim[\log_2 (n\varphi(n))]^{-\alpha},\,n\in\mathbb N,
\end{equation}
\begin{eqnarray}
\label{pq2}
{\rm (iii)}\quad\quad\quad\quad\quad\epsilon_n({\mathcal K})_X\lesssim 2^{-cn^{\alpha}},\,n\in\mathbb N,\,0<\alpha<1\quad\Rightarrow\quad d_n^{\gamma_n}({\mathcal K})_X\lesssim2^{-c(n\varphi(n))^{\alpha}}, \,n\in\mathbb N.
\end{eqnarray}
\end{theorem}
\noindent
{\bf Proof:} 
Indeed, it follows from  \cite[Theorem 3.3]{PW} that for any compact subset ${\mathcal K}\subset X$ of a Banach space $X$ and   any $n\geq 1$ we have that
	\begin{equation}
	\label{stat}
	 d^{2^k{\mathrm{rad}}({\mathcal K})}_{n}({\mathcal K})_X\leq \epsilon_{ kn}({\mathcal K})_X, \quad { k=1,2,\dots.}
	\end{equation}	
 We choose  $k=k(n)$ to be  such that 
$$
2^{k}{\rm rad}(\mathcal K)\leq 2^{\varphi(n)}<2^{k+1}{\rm rad}(\mathcal K)<2^{k+\ell},
$$
where ${\rm rad}(\mathcal K)<2^{\ell-1}$, then 
$ k>\varphi(n)-\ell>\lceil\frac{\varphi(n)}{2}\rceil$ for $n\geq n_0$, $n_0=n_0({\mathcal K})$  big enough.
Then
$$
d^{\gamma_n}_{n}({\mathcal K})_X\leq d^{2^k{\rm rad}({\mathcal K})}_{n}({\mathcal K})_X\leq \epsilon_{ kn}({\mathcal K})_X
\leq \epsilon_{ n\lceil{\frac{\varphi(n)}{2}}\rceil}({\mathcal K})_X, \quad \hbox{for}\quad n\geq n_0(\mathcal K),
$$
and therefore (\ref{gg}) holds.
Estimates (\ref{pq}), (\ref{pq1})  and (\ref{pq2}) follow from (\ref{gg}).
{Note that $n_0$ depends} only on ${\rm rad} (\cK)$ and on how fast $\varphi(n)$  grows to infinity.
\hfill $\Box$

\bigskip
Theorem \ref{cr1} and Theorem \ref{widthsfrombelownew}  can be combined in the next corollary. The latter can be viewed as a generalization of Corollary 
6.4 from \cite{PW}, which  covers the particular case $\varphi(n)=cn$.

\begin{cor}
\label{C11}
Let ${\mathcal K}\subset X$ be a compact set  of a Banach space $X$ and the function $\varphi:{\mathbb N}\to{\mathbb R}$ be such that   $\varphi(n)\geq c\log_2n$ for some fixed constant $c>0$. Let $d_n^{\gamma_n}({\mathcal K})_X$ be the  Lipschitz width  of $\mathcal K$ with 
 Lipschitz constant  $\gamma_n=2^{\varphi(n)}$. Then the following holds:
\begin{eqnarray*}
 \epsilon_n({\mathcal K})_X\asymp [\log_2 n]^\beta n^{-\alpha}, \, n\in \mathbb N, \,\alpha>0,\,\beta\in\mathbb R\quad&\Rightarrow& \quad d_n^{\gamma_n}({\mathcal K})_X\asymp[\log_2 (n\varphi(n))]^{\beta}[n\varphi(n)]^{-\alpha},\, n\in \mathbb N,\\
 \epsilon_n({\mathcal K}) _X\asymp[\log_2 n]^{-\alpha}\, n\in \mathbb N, \, \alpha>0\quad&\Rightarrow&\quad d_n^{\gamma_n}({\mathcal K})_X\asymp[\log_2 (n\varphi(n))]^{-\alpha},\, n\in \mathbb N.
\end{eqnarray*}
\end{cor}

\bigskip
 It follows from  Theorem 3.1 from \cite{PW}  that when  $\gamma$ is independent on $n$, i.e $\gamma=const$  
(this case is excluded in Corollary \ref{C11} because of the condition $\varphi(n)\geq c\log_2n$),  we do not have matching lower and upper bounds for $d_n^{\gamma}({\mathcal K})_X$ in the case when
$\epsilon_n({\mathcal K})_X\asymp [\log_2 n]^\beta n^{-\alpha}$, namely we have 
\begin{eqnarray*}
 \epsilon_n({\mathcal K})_X\lesssim [\log_2 n]^\beta n^{-\alpha},\, n\in \mathbb N, \, \alpha>0, \,\beta\in \mathbb R\quad&\Rightarrow& \quad d_n^{\gamma}({\mathcal K})_X\lesssim[\log_2 n]^{\beta}n^{-\alpha},,\, n\in \mathbb N,\\
 \epsilon_n({\mathcal K})_X\gtrsim [\log_2 n]^\beta n^{-\alpha},\, n\in \mathbb N, \, \alpha>0,\,\beta\in \mathbb R\quad&\Rightarrow&\quad d_n^{\gamma}({\mathcal K})_X\gtrsim[\log_2 n]^{\beta-\alpha}n^{-\alpha},\, n\in \mathbb N.
\end{eqnarray*}
It is still an open question whether the upper bound for $d_n^{\gamma}({\mathcal K})_X$ in this case can be improved to 
$d_n^{\gamma}({\mathcal K})_X\lesssim[\log_2 n]^{\beta-\alpha}n^{-\alpha}$. The following example, constructed in \cite{PW}, is in support of this conjecture. The  compact subset $\mathcal K(\sigma)\subset{\bf c_0}$ of the Banach space ${\bf c_0}$  of all sequences that converge to 0, equipped with the
$\ell_\infty$  norm, defined as
$$
\mathcal K(\sigma):=\{\sigma_je_j\}_{j=1}^\infty\cup \{0\},\quad \sigma_j:=(\log_2(j+1))^{-1},
$$
where  $(e_j)_{j=1}^\infty$  are the
standard basis in ${\bf c_0}$ has entropy numbers 
$\epsilon_n(\mathcal K(\sigma))_{\bf c_0}\asymp n^{-1}$ and Lipschitz width 
$d_n^{\gamma}({\mathcal K(\sigma)})_{\bf c_0}\asymp n^{-1}[\log_2(n+1)]^{-1}$, $\gamma=const$. 
\begin{cor}
It follows from  Lemma {\rm \ref{L2a}} { and Corollary {\rm \ref{C11}} that  if a compact subset $\mathcal K\subset X$ of a Banach space $X$ has entropy numbers
$\epsilon_n({\mathcal K})_X\asymp [\log_2 n]^\beta n^{-\alpha}$ or $\epsilon_n({\mathcal K}) _X\asymp[\log_2 n]^{-\alpha}$, then for any $c_1,\tilde c>0$ and $a,a_1>0$, we have}
$$
 d_n^{2^{\tilde cn}}(\mathcal K)_X\asymp d^{2^{c_1n}}(\mathcal K,\ell^n_\infty)_X,\quad d_n^{2^{an\log_2n}}(\mathcal K)_X\asymp d^{2^{a_1n\log_2n}}(\mathcal K,\ell^n_\infty)_X,
$$
where the constants in $\asymp$ depend on $\tilde c$ and $c_1$,  or $a$ and $a_1$, respectively.
\end{cor}
\noindent
{\bf Proof:} We consider the case when $\gamma_n=2^{\tilde cn}$  (the case $\gamma_n=2^{an\log_2n}$ is similar). 
 We take  any $c_1>0$ and fix  $c\leq c_1$. Then, for $n$ big enough we have $c_1n\geq cn+\log_2(2n+1)$ and it follows  from the monotonicity of the fixed Lipschitz widths and  Lemma \ref{L2a}  that
$$
d_n^{2^{c_1n}}(\mathcal K)_X\leq d^{2^{c_1n}}(\mathcal K,\ell^n_\infty)_X \leq d^{(2n+1)2^{cn}}(\mathcal K,\ell^n_\infty)_X\leq d_n^{2^{cn}}(\mathcal K)_X.
$$
On the other hand,  $d_n^{2^{cn}}(\mathcal K)_X\asymp d^{2^{c_1n}}(\mathcal K)_X\asymp d^{2^{\tilde c n}}(\mathcal K)_X$, see Corollary \ref{C11}, and therefore
$$
d^{2^{c_1n}}(\mathcal K,\ell^n_\infty)_X\asymp d_n^{2^{cn}}(\mathcal K)_X\asymp d_n^{2^{\tilde cn}}(\mathcal K)_X.
$$
 \hfill $\square$}


\section{Bounds for the entropy numbers via  Lipschitz widths and the error of neural network  approximation }
\label{UB}
While Corollary \ref{C11} deduces what  the behavior of the Lipschitz widths is  if we know the behavior of the entropy numbers of a class $\mathcal K$, we can ask the inverse question. Namely, what does the 
asymptotic behavior of the Lipschitz widths tell us about the entropy numbers of $\mathcal K$? 
Any theorem that deduces the behavior of the entropy numbers of a class $\mathcal K$ from its Lipschitz width  $d^{\gamma_n}_n(\mathcal K)_X$  could shed a light on the entropy numbers of 
the classes of functions that are well approximated via deep or shallow NNs.   

\subsection{ Upper bounds for entropy numbers via Lipschitz widths}
We start with a lemma that is an extension of Lemma 3.7 from \cite{PW}.

\begin{lemma} 
\label{LL1}
If  $\cK \subset X$ is a compact set, $\gamma_n=2^{\varphi(n)}$, and  $d^{\gamma_n}_n(\cK)_X< \epsilon/2$, then 
we have the following bound for the entropy number 
$$
\epsilon_r(\cK)_X\leq \epsilon, \quad \hbox{where}\quad r\geq\lceil n(\varphi(n)+ \log_2(6/\epsilon))\rceil.
$$
In particular:

\hskip -0.2in {\rm (i)}  Let  $d_n^{\gamma_n}(\cK)_X=0$ for  some  $n\in \mathbb N$. Then
 for any  $k\in \mathbb N$    such that    $k>\varphi(n)$, we have
\begin{equation}
\label{klkl}
\epsilon_{nk}(\mathcal K)_X\leq 6\cdot 2^{\varphi(n)-k}= 6\cdot \gamma_n 2^{-k}.
\end{equation}

\hskip -0.2in {\rm (ii)} Let  $\gamma_n= 2^{cn^p[\log_2n]^q}$ for some $p\geq 0$ and $q\in \mathbb R$.
\begin{itemize}
\item  If for some $\alpha>0$, $\beta\in \mathbb R$, we have $0<d^{\gamma_n}_n(\cK)_X\lesssim [\log_2n]^\beta n^{-\alpha}$, $n\in\mathbb N$, 
 then
 \begin{itemize}
\item  when $p>0$ and $q\in \R$, or  $p=0$ and $q\geq 1$, we have 
$$ 
\epsilon_n(\cK)_X\lesssim n^{-\alpha/(1+p)} [\log_2 n]^{\beta+\frac{\alpha q}{1+p}}, \quad n\in\mathbb N;
$$
\item when   $p=0$ and $ q<1$, we have  
$$
\epsilon_n(\cK)_X\lesssim n^{-\alpha} [\log_2 n]^{\alpha+ \beta}, \quad n\in\mathbb N;
$$
\end{itemize}
\item  If  for some $c>0$, we have $0<d^{\gamma_n}_n(\cK)_X\lesssim 2^{-cn}$, $n\in\mathbb N$,  then
\begin{itemize}
\item  when $0\leq p<1$ and $q\in \mathbb R$, or  $p=1$ and $q\leq 0$, we have 
$$ 
\epsilon_n(\cK)_X\lesssim 2^{-c\sqrt{n}}, \quad n\in\mathbb N;
$$
\item when   $p>1$ and $ q\in \mathbb R$ or $p=1$, $q>0$, we have  
$$
\epsilon_n(\cK)_X\lesssim 2^{-c_1n^{1/(p+1)}[\log_2n]^{-q/(p+1)}}, \quad n\in\mathbb N.
$$
\end{itemize}
\end{itemize}
\end{lemma}
\noindent
{\bf Proof:} We provide the proof in the Appendix.
\hfill $\Box$

\begin{remark}
\label{newcor}
It follows from \eref{main} that
 in  Lemma {\rm \ref {LL1} } we can take $\epsilon=6E(\cK,\Sigma_{n,\sigma}(w(n)))_X>0$ 
 or $\epsilon=6E(\cK,\Xi_{W,\sigma}(w(W)))_X>0$ and  obtain that
$$
\epsilon_r(\cK)_X\leq 6E(\cK,\Sigma_{n,\sigma}(w(n)))_X, \quad  r\geq\lceil cn^2(1+\log_2w(n))- \log_2(E(\cK,\Sigma_{n,\sigma}(w(n)))_X))\rceil,
$$
 or 
$$
\epsilon_r(\cK)_X\leq 6E(\cK,\Xi_{W,\sigma}(w(W)))_X, \quad  r \geq\lceil cW(\log_2W+\log_2w(W))- \log_2(E(\cK,\Xi_{W,\sigma}(w(W)))_X))\rceil,
$$
The case $E(\cK,\Sigma_{n,\sigma}(w(n)))_X=0$  or $E(\cK,\Xi_{W,\sigma}(w(W)))_X=0$ is the same as when  $d^{\gamma_n}_n(\cK)_X=0$,
 provided $C([0,1]^d)$ is continuously embedded in $X$.
 \end{remark}

 \subsection{ Upper bounds for entropy numbers via DNN approximation rates.}
 Let us  now  consider a compact set  $\mathcal K\subset X$,  where $C([0,1]^d)$ is continuously embedded in $X$. The next  corollary is a direct consequence of Lemma \ref{LL1}.
 \begin{cor}
\label{NNA}
The following holds:

\hskip -0.2in
 {\rm (i)} 
Let $E(\cK,\Sigma_{n,\sigma}(w(n)))_X\lesssim[\log_2n]^\beta n^{-\alpha}$, $n\in \mathbb N$ for some $\alpha>0$  and  $\beta\in \mathbb R$.
\begin{itemize}
\item When  $w(n)=Cn^\delta$, $C>0$, $\delta>0$, we have 
\begin{equation}
\label{tt1a}
\epsilon_n(\mathcal K)_X\lesssim n^{-\frac{\alpha}{2}}[\log_2n]^{\beta+\frac{\alpha}{2}}, \quad n\in\mathbb N.
\end{equation}
\item When $w(n)=C2^{cn^\nu}$, $C,c>0$,  $\nu\geq 0$, we have  
\begin{equation}
\label{tta}
\epsilon_n(\mathcal K)_X\lesssim n^{-\frac{\alpha}{\nu+2}}[\log_2n]^{\beta}, \quad n\in\mathbb N.
\end{equation}
\end{itemize}

\hskip -0.2in {\rm (ii)}  Let  $E(\cK,\Sigma_{n,\sigma}(w(n)))_X\lesssim  2^{-cn}$, $n\in \mathbb N$ for some $c>0$. 
\begin{itemize}
\item When  $w(n)=Cn^\delta$, with $C,\delta>0$, we have 
\begin{equation}
\label{tt1}
\epsilon_n(\mathcal K)_X\lesssim 2^{-{c_1}\sqrt{n}/\sqrt{\log_2n}}, \quad n\in\mathbb N.
\end{equation}
\item  When $w(n)=C2^{cn^\nu}$, $C,c>0$,  $\nu\geq 0$,  we have 
\begin{equation}
\label{tt}
\epsilon_n(\mathcal K)_X\lesssim 2^{-c_1n^{1/(\nu+2)}}, \quad n\in\mathbb N.
\end{equation}
\end{itemize}
\end{cor}
\noindent
{\bf Proof:} The proof follows directly from Lemma  \ref{LL1} and inequality (\ref{main}). Note that in this lemma we require that  
$d_n^{\gamma_n}(\cK)_X>0$.  However, it could happen that  for some $n$ we have $d_n^{\gamma_n}(\cK)_X=0$. Then, we proceed as follows: 
\begin{itemize}
\item   When $w(n)=Cn^\delta$, we have that, see Table 1,  $\varphi(n)=c'n\log_2n$, and therefore we can use (\ref{klkl}) for $k=2c'n\log_2n$ to derive that
$$
\epsilon_{2c'n^2\log_2n}(\mathcal K)_X\leq 6\cdot 2^{-c'n\log_2n}.
$$
For any $m$ large enough, we can find $n=n(m)$, such that
$$
2c'n^2\log_2n\leq m<2c'(n+1)^2\log_2(n+1)<c_1n^2\log_2n,
$$
and therefore
$$
\epsilon_{m}(\mathcal K)_X\leq \epsilon_{2c'n^2\log_2n}(\mathcal K)_X\leq 6\cdot 2^{-c'n\log_2n}.
$$
Note that
$$
-c'n\log_2n\leq -\frac{c'}{c_1}\frac{m}{n}, \quad n\lesssim \frac{\sqrt{m}}{\sqrt{\log_2 n}}, \quad \log_2n\asymp \log_2m,\quad \Rightarrow   -n\log_2n\lesssim -\sqrt{m}\sqrt{\log_2 m},
$$
and thus
$$
\epsilon_{m}(\mathcal K)_X\leq 6\cdot 2^{-\tilde c\sqrt{m}\sqrt{\log_2 m}},
$$
which agrees with (\ref{tt1a}) and  \eref{tt1}.
\item  When  $w(n)=C2^{cn^\nu}$, we have that, see Table 1,  $\varphi(n)=c'n^{\nu+1}$, and  therefore we can use (\ref{klkl})  for $k=2c'n^{\nu+1}$, to obtain that 
$$
\epsilon_{2c'n^{\nu+2}}(\mathcal K)_X\leq 6\cdot 2^{-c'n^{\nu+1}}\quad \Rightarrow \quad \epsilon_{n}(\mathcal K)_X\leq C\cdot 2^{-c_1\sqrt{n}}.
$$
For any $m$ large enough, we can find $n=n(m)$, such that
$$
2c'n^{\nu+2}\leq m<2c'(n+1)^{\nu+2}<c_1n^{\nu+2}\quad \Rightarrow\quad m^{1/(\nu+2)}\asymp n,
$$
and therefore
$$
\epsilon_{m}(\mathcal K)_X\leq \epsilon_{2c'n^{\nu+2}}(\mathcal K)_X\leq 6\cdot 2^{-c'n^{\nu+1}}\lesssim 2^{-\tilde cm^{\frac{\nu+1}{\nu+2}}},
$$
which agrees with (\ref{tta}) and (\ref{tt}).
\end{itemize}
\hfill $\Box$

 \subsection{Approximation classes for DNNs}
 \label{Apclasses}

Recall that if  $f\in X$ and $A$ is a subset of the Banach space $X$, the distance between $f$ and $A$ is defined as ${\rm dist}(f,A)_{X}:=\inf_{g\in A}\|f-g\|_X$. Clearly, we have that
$$
E(f,\Sigma_{n,\sigma}(w(n)))_X={\rm dist}(f,\Phi_\sigma(B_{\ell_\infty}^{\tilde n}(w(n)))_X,
$$ 
where the sets $\Phi_\sigma(B_{\ell_\infty}^{\tilde n}(w(n)))\subset X$ are compact with respect to the uniform norm $C(\Omega)$, see Theorem \ref{NN}.

Let ${\bf \xi}:=(\xi_n)_{n=1}^\infty$ be a sequence of non-negative numbers such that  $\displaystyle{\inf_n\xi_n=0}$ (in particular, we can have $\lim_{n\to\infty}\xi_n=0$). We  denote by ${\mathcal N}_{{\bf \xi},\sigma}(w)$ the set
of functions that are approximated by functions from  $\Sigma_{n,\sigma}(w(n))$  with accuracy $\xi_n$  for every $n\geq 0$. More precisely,
$$
{\mathcal N}_{{\bf \xi},\sigma}(w):= \{f\in X:\,  \,E(f,\Sigma_{n,\sigma}(w(n)))_X\leq \xi_n, \,\forall n\geq 0\},
$$
 which can be written equivalently as
$$
{\mathcal N}_{{\bf \xi},\sigma}(w)=\bigcap_{n=0}^\infty V_n(\xi), \quad V_n(\xi):=\{f\in X:\,{\rm dist}(f,\Phi_\sigma(B_{\ell_\infty}^{\tilde n}(w(n)))_X\leq \xi_n\}.
$$
Then,  if $X$ is such that $C(\Omega)$ is continuously embedded in $X$, we can apply Lemma \ref{Pr}  (see the Appendix)  to obtain that ${\mathcal N}_{{\bf \xi},\sigma}(w)$ is a (possibly empty) compact subset of $X$.
 In what follows, we show that there are choices of sequences $\xi$ and DNNs with bounds $w$ on their parameters  for which the compact sets ${\mathcal N}_{{\bf \xi},\sigma}(w)\neq \emptyset$.

\begin{remark}
 According to Remark {\rm \ref{Sremark}}, the set   $\Sigma_{n,\sigma}^{s}(w(n))$ of outputs of a DNN where at each layer one uses $\bar \sigma=(\sigma_0,\sigma, \ldots,\sigma,\sigma_0)$ with $\sigma_0(t)=t$,  
and $\sigma=$ReLU, see  \eref{NN1}, satisfy Theorem {\rm \ref{NN}. Therefore, all theory developed so far holds for $\Sigma_{n,\sigma}^{s}(w(n))$. 
}
 
Let us now consider the case when $\Omega=[0,1]$ and $\sigma=$ReLU. If we denote by $H$ the hat function $ H(t)=2(t-0)_+-4(t-\frac{1}{2})_+$ and by $H^{\circ k}$ this function composed with itself $k$ times, then,
 for properly selected $w(n)$, we have the inclusion, see  Figure {\rm 6.1},
$$
\{\psi_n:=\sum_{k=1}^nc_kH^{\circ k}\}\subset \Sigma_{n,\sigma}^{s}(w(n)).
$$
\begin{figure}[h]
 \label{F1}
 \begin{center}
\includegraphics[width=4in]{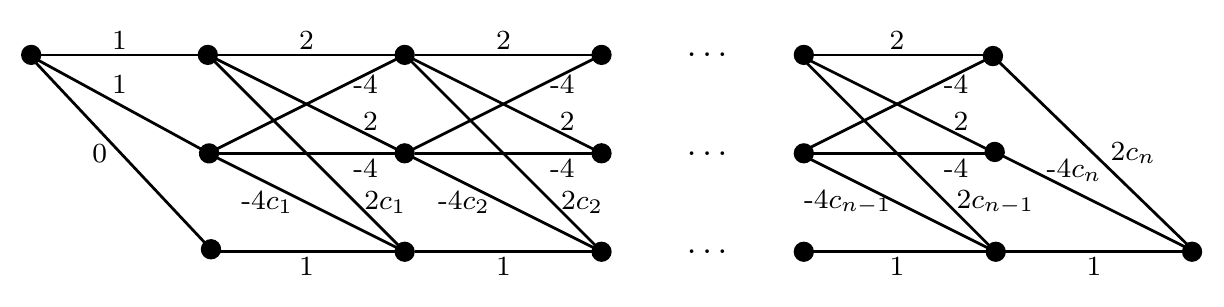}
\end{center}
\caption{Computational graph for $\psi_n$}
\end{figure}

\noindent
Recall that  the  set $\cal T$ defined as 
$$
 {\cal T}:=\left\{f:\,f=\sum_{k=1}^\infty c_kH^{\circ k}, \,\sum_{k=1}^\infty|c_k|<\infty\right\},
$$
is called the  Takagi class,  see {\rm\cite{Ta}}. Clearly
 $
 {\cal R}:=\{f_\lambda=\sum_{k=1}^\infty \lambda^{-k}H^{\circ k}, \quad |\lambda|>1\}\subset \mathcal T,
$
and it is a well known fact that   the Takagi function $T=f_2$ and  that $t(1-t)=f_4(t)$, see {\rm \cite{Ta}}.

 Let us observe that the elements of $\cal R$  can be approximated with exponential accuracy by outputs of $\Sigma_{n,\sigma}^{s}(w(n))$ with $w(n)=C$.
Therefore,  in this case the set 
${\mathcal N}_{{\bf \tilde \xi},\sigma}(w)$, with  $\tilde \xi=(C2^{-cn})_{n=1}^\infty$ is non-empty, and 
 for every $\eta=(\eta_n)_{n=1}^\infty$ with the property that $\inf_n\eta_n=0$ and $C2^{-cn}\leq \eta_n$ for all $n\geq 0$, we have
$$
\emptyset \ne {\mathcal N}_{{\bf \tilde \xi},\sigma}(w)\subset {\mathcal N}_{{\bf \eta},\sigma}(w).
$$
\end{remark}

\bigskip
Now, let us return to the relation between the error $E({\mathcal N}_{{\bf \xi},\sigma}(w),\Sigma_{n,\sigma}(w(n)))_X$ and the Lipschitz width. It follows from the definition of 
${\mathcal N}_{{\bf \xi},\sigma}(w)$ that for every $n\geq 0$, 
$E({\mathcal N}_{{\bf \xi},\sigma}(w),\Sigma_{n,\sigma}(w(n)))_X\leq \xi_n$, and therefore  (\ref{main}) gives that 
\begin{equation}
\label{main2}
d^{\gamma_n}_n({\mathcal N}_{\xi,\sigma}(w))_X\leq \xi_n, \quad \hbox{with}\quad \gamma_n=2^{cn(1+\log_2w(n))}.
\end{equation}
One can now apply Lemma \ref{LL1}  and derive estimates for the entropy numbers of the compact set  ${\mathcal N}_{\xi,\sigma}(w)$. 
 Such estimates can be viewed as inverse theorems for DNN approximation. 
 
 To simplify the presentation, let us consider 
 the following special sequences $\xi_n=C[\log_2n]^\beta n^{-\alpha}$ and 
$\xi_n=C2^{-cn}$, and 
DNNs with depth $n$, ReLU or Lipschitz sigmoidal activation function and bounds on the NN parameters $w(n)=Cn^\delta$, $\delta\geq 0$ and $w(n)=C2^{cn^\nu}$, $\nu\geq 0$, see Table 1. 
Note that the case $\nu=0$ covers the case 
$w(n)=C$. 
 We apply Corollary \ref{NNA} to the class $\mathcal K={\mathcal N}_{{\bf \xi},\sigma}(w)$ and obtain the following result. 
\begin{cor}
\label{NNAnew}
The entropy numbers of the compact set  ${\mathcal N}_{\xi,\sigma}(w)$ that consists of  all functions  approximated in the norm of $X$ with accuracy $\xi_n$ by the outputs of  
DNN with depth $n$, ReLU or Lipschitz sigmoidal activation function and bounds on the NN parameters 
$w$ satisfy the following inequalities for the listed special choices of $w$ and sequences $(\xi)$:
\begin{itemize}
\item  If $C,c,\alpha,\delta>0$, $\nu\geq 0$, and $\beta\in \mathbb R$, we have 
\begin{equation}
\nonumber
\epsilon_n({\mathcal N}_{\xi,\sigma}(Cn^\delta))_X\lesssim n^{-\frac{\alpha}{2}}[\log_2n]^{\beta+\frac{\alpha}{2}}, \quad n\in\mathbb N, \quad \hbox{where}\quad \xi=(\xi_n)=(C[\log_2n]^\beta n^{-\alpha}),
\end{equation}
\begin{equation}
\nonumber
\epsilon_n({\mathcal N}_{\xi,\sigma}(C2^{cn^\nu}))_X\lesssim n^{-\frac{\alpha}{\nu+2}}[\log_2n]^{\beta}, \quad n\in\mathbb N, \quad \hbox{where}\quad\xi=(\xi_n)=(C[\log_2n]^\beta n^{-\alpha}).
\end{equation}
\item    If $C,c,\alpha,\delta>0$, $\nu\geq 0$, and $\beta\in \mathbb R$, we have 
\begin{equation}
\nonumber
\epsilon_n({\mathcal N}_{\xi,\sigma}(Cn^\delta))_X\lesssim 2^{-{c_1}\sqrt{n}/\sqrt{\log_2n}}, \quad n\in\mathbb N, \quad \hbox{where}\quad\xi=(\xi_n)=(C2^{-cn}),
\end{equation}
\begin{equation}
\nonumber
\epsilon_n({\mathcal N}_{\xi,\sigma}(C2^{cn^\nu}))_X\lesssim 2^{-c_1n^{1/(\nu+2)}}, \quad n\in\mathbb N, \quad \hbox{where}\quad\xi=(\xi_n)=(C2^{-cn}).
\end{equation}
\end{itemize}
\end{cor}

\begin{remark}
The above  estimates hold  also for the sets $\lambda {\mathcal N}_{{\bf \xi},\sigma}(w)=\{\lambda f:\,f\in {\mathcal N}_{{\bf \xi},\sigma}(w)\}$ with $\lambda>1$, 
 where the constants involved depend on $\lambda$. Indeed, the fact that $f\in V_n(\xi)$ implies the inequality 
$$
{\rm dist}(\lambda f,\Phi_\sigma(B_{\ell_\infty}^{\tilde n}(\lambda w(n))))_X\leq \lambda \xi_n,
$$
since $\lambda \Phi_\sigma(B_{\ell_\infty}^{\tilde n}( w(n)))\subset \Phi_\sigma(B_{\ell_\infty}^{\tilde n}(\lambda w(n)))$. Then,
according to  \eref{main} and the monotonicity of Lipschitz widths, we have
$$
d^{\gamma'_n}_n(\lambda {\mathcal N}_{{\bf \xi},\sigma}(w))_X\leq \lambda\xi_n, \quad \hbox{with}\quad \gamma'_n:=2^{c(\lambda)n(1+\log_2w(n))}>2^{cn(1+\log_2(\lambda w(n)))},
$$
and we can apply Lemma {\rm \ref{LL1}} or Corollary {\rm \ref{NNA}}.
\end{remark}

\subsection{ Upper bounds for entropy numbers via SNN approximation rates}
 \label{Spclasses}
 In this section we study classes $\mathcal K$ for which $E(\mathcal K,\Xi_{W,\sigma}(w(W)))_X\leq \xi_W$ for general sequences $(\xi_W)$ with non-negative terms
 and $X$ is a Banach space such that $C([0,1]^d)\subset X$ is continuously embedded in $X$.
Since  $\Xi_{W,\sigma}(w(W))\subset \Xi_{W+1,\sigma}(w(W))$, we require 
 $(\xi_W)$ to be non-increasing sequence with 
 $$
 \lim_{W\to\infty}\xi_W=0.
 $$
 For example, we  consider 
the sequences $\xi_W=C[\log_2W]^\beta W^{-\alpha}$ and 
$\xi_W=C2^{-cW}$, and 
 SNNs with  ReLU or Lipschitz sigmoidal activation function and bounds on the NN parameters $w(W)=CW^\delta$, $\delta\geq 0$ 
  and $w(W)=C2^{cW^\nu}$, $\nu>0$. Clearly  when  $\delta=0$ we have
$w(W)=C$. It follows from  Table 2  that we have
\begin{eqnarray*}
d_W^{\gamma_W}(\mathcal K)_X&\leq&  \xi_W, \quad \gamma_W=2^{c'\log_2W}, \quad \delta\geq0,\\
d_W^{\gamma_W}({\mathcal K})_X&\leq & \xi_W, \quad \gamma_W=2^{c'W^{\nu}},
\end{eqnarray*}
and we can apply  Lemma   \ref{LL1}  with $p=0$, $q=1$ (when $w(W)=CW^\delta$),  
 and $p=\nu$, $q=0$ (when $w(W)=C2^{cW^\nu}$). 
More precisely, the following corollary holds.
 \begin{cor}
\label{NNAs}
Let ${\mathcal K}\subset X$ be a compact subset of a Banach space $X$,  where $X$ is such that $C([0,1]^d)\subset X$ is continuously embedded in $X$.
 Then the following holds:

\hskip -0.2in
 {\rm (i)} 
Let $E(\mathcal K,\Xi_{W,\sigma}(w(W)))_X\lesssim [\log_2W]^\beta W^{-\alpha}$, $W\in \mathbb N$, for some  $\alpha>0$  and  $\beta\in \mathbb R$.
\begin{itemize}
\item When  $w(W)=CW^\delta$, $C>0$, $\delta\geq 0$, we have 
\begin{equation}
\label{tt1asnew}
\epsilon_n({\mathcal K})_X\lesssim n^{-\alpha}[\log_2n]^{\beta+\alpha}, \quad n\in\mathbb N.
\end{equation}
\item When $w(W)=C2^{cW^\nu}$, $C,c>0$,  $\nu> 0$, we have  
\begin{equation}
\label{ttasnew}
\epsilon_n({\mathcal K})_X\lesssim n^{-\frac{\alpha}{\nu+1}}[\log_2n]^{\beta}, \quad n\in\mathbb N.
\end{equation}
\end{itemize}

\hskip -0.2in {\rm (ii)}  Let  $E(\mathcal K,\Xi_{W,\sigma}(w(W)))_X\lesssim 2^{-cW}$, $W\in \mathbb N$,  for some $c>0$. 
\begin{itemize}
\item When  $w(W)=CW^\delta$, with $C>0$, $\delta\geq 0$, we have 
\begin{equation}
\label{tt1snew}
\epsilon_n({\mathcal K})_X\lesssim 2^{-c_1\sqrt{n}}, \quad n\in\mathbb N.
\end{equation}
\item  When $w(W)=C2^{cW^\nu}$, $C,c,\nu>0$,  we have 
\begin{equation}
\label{ttsnew}
\epsilon_n({\mathcal K})_X\lesssim 2^{-c_1n^{1/(\nu+1)}}, \quad n\in\mathbb N.
\end{equation}
\end{itemize}
\end{cor}
\noindent
{\bf Proof:} The statement follows  from Lemma  \ref{LL1} and  (\ref{mains}). Lemma  \ref{LL1}  requires that  
$d_W^{\gamma_W}(\cK)_X>0$. If   for some $W$ we have $d_W^{\gamma_W}(\cK)_X=0$, we proceed as follows:
\begin{itemize}
\item   When $w(W)=CW^\delta$, we have that, see Table 2,  $\varphi(W)=c'\log_2W$, and therefore we can use (\ref{klkl}) for $k=W+c'\log_2W$ to derive that
$$
\epsilon_{W(W+c'\log_2W)}({\mathcal K})_X\leq 6\cdot 2^{-W}.
$$
For any $m$ large enough, we can find $W=W(m)$, such that
$$
W^2<W(W+c'\log_2W)\leq m<(W+1)(W+1+c'\log_2(W+1))<c_1W^2,
$$
and therefore
\begin{equation}
\label{mi}
\epsilon_{m}({\mathcal K})_X\leq \epsilon_{W(W+c'\log_2W)}({\mathcal K})_X\leq 6\cdot 2^{-W}< 6\cdot 2^{-\tilde c\sqrt{m}},
\end{equation}
which agrees with (\ref{tt1asnew}) and  \eref{tt1snew}.
\item  When  $w(W)=C2^{cW^\nu}$, we have  $\varphi(W)=c'W^{\nu}$ and   we can use (\ref{klkl})  for $k=c'W^{\nu}+W^{\nu+1}=W^{\nu}(c'+W)$, to obtain that 
$$
\epsilon_{c'W^{\nu}(c'+W)}({\mathcal K})_X\leq 6\cdot 2^{-W^{\nu+1}}.
$$
For any $m$ large enough, we can find $W=W(m)$, such that
$$
W^{\nu+1}<W^{\nu}(c'+W)\leq m<(W+1)^{\nu}(c'+W+1)\leq c_1W^{\nu+1}
$$
and therefore
$$
\epsilon_{m}({\mathcal K})_X\leq \epsilon_{c'W^{\nu}(c'+W)}({\mathcal K})_X\leq 6\cdot 2^{-'W^{\nu+1}}\lesssim 2^{-\tilde cm}
$$
which agrees with (\ref{ttasnew}) and (\ref{ttsnew}).
\end{itemize}
The proof is completed.
\hfill $\Box$

 \begin{remark}
Similar statement as Corollary {\rm \ref{NNAs}} was proven in Theorem {\rm 10} from {\rm \cite {JS}} in the case $\sigma_W=CW^{-\alpha}$  for classes 
$$
{\mathcal K}=\overline{\{f=\sum_{j=1}^nc_jh_j:\,h_j\in\mathbb D, \sum_{j=1}^n|a_j|\leq 1\}},
$$
 which are the closure of the convex, symmetric hull of
 dictionaries $\mathbb D$ satisfying specific properties and for SNNs with certain activation functions.
\end{remark}

 \subsection{Approximation classes for SNNs}
 \label{ACSSN}
 We consider Banach spaces $X$  such that $C([0,1]^d)\subset X$ is continuously embedded in $X$.
We   denote by ${\mathcal A}_{{\bf \xi},\sigma}(w)\subset X$ the approximation class
$$
{\mathcal A}_{{\bf \xi},\sigma}(w):= \{f\in X:\,  \,E(f,\Xi_{W,\sigma}(w(W)))_X\leq \xi_W, \,\forall W\geq 0\},
$$
or equivalently written as
$$
{\mathcal A}_{{\bf \xi},\sigma}(w)=\bigcap_{W=0}^\infty {\mathcal V}_W(\xi), \quad {\mathcal V}_W(\xi):=\{f\in X:\,{\rm dist}(f,\Psi_\sigma(B_{\ell_\infty}^{\tilde W}(w(W)))_X\leq \xi_W\}.
$$
As in the case of DNNs,  ${\mathcal A}_{{\bf \xi},\sigma}(w)$ is a compact subset of $X$, see Lemma \ref{Pr},  and, see 
  (\ref{mains}), its Lipschitz widths satisfy the inequalities
\begin{equation}
\label{main2s}
d^{\gamma_W}_W({\mathcal A}_{\xi,\sigma}(w))_X\leq \xi_W, \quad \hbox{with}\quad \gamma_W=2^{c(\log_2W+\log_2w(W))},\quad W\in \mathbb N.
\end{equation}
We next   apply Corollary  \ref{NNAs}  in the case   ${\mathcal K}={\mathcal A}_{\xi,\sigma}(w)$
and derive the following statement.

\begin{cor}
\label{NNAsnew}
The entropy numbers of the approximation class ${\mathcal A}_{\xi,\sigma}(w)$ that consists of  all functions  approximated in the norm of $X$ with accuracy $\xi_W$ by the outputs of  
a SNN with width $W$, ReLU or Lipschitz sigmoidal activation function and bounds on the NN parameters 
$w(W)=CW^\delta$, $\delta\geq 0$ or  $w(W)=C2^{cW^\nu}$, $C,c>0$, $\nu> 0$, satisfy the following inequalities:

\hskip -0.2in
 {\rm (i)} 
Let $\xi_W=C[\log_2W]^\beta W^{-\alpha}$ for some $C>0$, $\alpha>0$  and  $\beta\in \mathbb R$.
\begin{itemize}
\item If $C,C',\alpha,\nu>0$, $\delta\geq 0$, $\beta\in \mathbb R$, we have 
\begin{equation}
\nonumber
\epsilon_n({\mathcal A}_{\xi,\sigma}(CW^\delta))_X\lesssim n^{-\alpha}[\log_2n]^{\beta+\alpha}, \quad n\in\mathbb N, \quad \hbox{where}\quad \xi=(\xi_W)=(C'[\log_2W]^\beta W^{-\alpha}),
\end{equation}
\begin{equation}
\nonumber
\epsilon_n({\mathcal A}_{\xi,\sigma}(C2^{cW^\nu}))_X\lesssim n^{-\frac{\alpha}{\nu+1}}[\log_2n]^{\beta}, \quad n\in\mathbb N,\quad  \hbox{where}\quad \xi=(\xi_W)=(C'[\log_2W]^\beta W^{-\alpha}).
\end{equation}
\item If   $C,C',c,\nu>0$, $\delta\geq 0$, we have 
\begin{equation}
\nonumber
\epsilon_n({\mathcal A}_{\xi,\sigma}(CW^\delta))_X\lesssim 2^{-c_1\sqrt{n}}, \quad n\in\mathbb N,\quad  \hbox{where}\quad \xi=(\xi_W)=(C'2^{-cW}),
\end{equation}
\begin{equation}
\nonumber
\epsilon_n({\mathcal A}_{\xi,\sigma}(C2^{cW^\nu}))_X\lesssim 2^{-c_1n^{1/(\nu+1)}}, \quad n\in\mathbb N,\quad  \hbox{where}\quad \xi=(\xi_W)=(C'2^{-cW}).
\end{equation}
\end{itemize}
\end{cor}


\subsection{Further study of  Lipschitz widths with large Lipschitz constants, part II}
\label{p2a}
In this section, we study the relationship between  the Lipschitz widths of a compact class $\cK$ and  its entropy numbers. 
We start with the following refined version of Lemma \ref{LL1}.

\begin{lemma} 
\label{newlemma}
Let $\gamma_n=2^{\varphi(n)}$  with  $\varphi:[1,\infty)\to \mathbb R$ an increasing non-negative function such that $\varphi(n)\to\infty$ as $n\to\infty$
and $d_n^{\gamma_n}(\mathcal K)_X\geq c_02^{-\varphi(n)}$ for some $c_0>0$. Then, 
 for $n$ sufficiently large, we have that
\begin{equation}
\label{rest}
d_{8n}^{\gamma_{8n}}(\cK)_X\leq \epsilon_{8 n\lceil\frac{\varphi(8n)}{2}\rceil}(\cK)_X\leq 3d_n^{\gamma_n}(\cK)_X.
\end{equation}
In particular, if
$ d_n^{\gamma_n}(\cK)_X\asymp  [\log_2(n\varphi(n))]^\beta [n\varphi(n)]^{-\alpha}$, and 
\begin{itemize}
\item if   the function $\varphi$ is such that there is a constant $c_1>1$ for which $\displaystyle{\sup_{n\in \mathbb N} \frac{\varphi(c_1n)}{\varphi(n)}<\infty}$ then  
$$
\epsilon_{m}(\cK)_X\asymp  [\log_2 m]^\beta m^{-\alpha}, \quad m\in \mathbb N.
$$
\item if for every $c>1$,   $\displaystyle{\sup_{n\in \mathbb N} \frac{\varphi(cn)}{\varphi(n)}= \infty}$,  then the   lower and upper bound in  {\rm (\ref{rest}) } are asymptotically different in the sense that
$$
 \frac{d_n^{\gamma_n}(\cK)_X}{d_{8n}^{\gamma_{8n}}(\cK)_X}\gtrsim \begin{cases}\left[\frac{\varphi(8n)}{\varphi(n)}\right]^\alpha
\left[ \log_2 \frac{\varphi(8n)}{\varphi(n)}\right]^{-\beta}, \quad \beta>0,\\ \\
 \left[\frac{\varphi(8n)}{\varphi(n)}\right]^\alpha, \quad\quad\quad\quad\quad\quad\quad\beta\leq 0.
 \end{cases}
$$
\end{itemize}
\end{lemma}
\noindent
{\bf Proof:} Following the proof of Lemma \ref{LL1}  and using the fact that  $d^{\gamma_n}_n(\cK)_X\geq c_02^{-\varphi(n)}$, we obtain 
for   the choice $\epsilon=3d^{\gamma_n}_n(\cK)_X>0$  that the entropy numbers $\epsilon_r(\cK)_X\leq 3d^{\gamma_n}_n(\cK)_X$, with
\be
\label{gh}
r=\lceil n(\varphi(n)+\log_2(2[d^{\gamma_n}_n(\cK)_X]^{-1}))\rceil\leq \lceil n(2\varphi(n)+\tilde c)\rceil\leq  \lceil 4n\varphi(n)\rceil,
\ee
provided  $n$ is sufficiently large, and thus 
$$
\epsilon_{\lceil 4n\varphi(n)\rceil}(\cK)_X\leq 3d^{\gamma_n}_n(\cK)_X.
$$
On the other hand,  Theorem \ref{cr1}, gives that
$$
d_{8n}^{\gamma_{8n}}(\cK)_X\leq \epsilon_{ 8n\lceil\frac{\varphi(8n)}{2}\rceil}(\cK)_X.
$$
We derive (\ref{rest}) from  the monotonicity of the entropy numbers, the latter two inequalities and 
  the fact that $\varphi$, as an increasing non-negative function, satisfies the condition  $8n\lceil\frac{\varphi(8n)}{2}\rceil\geq  \lceil 4n\varphi(n)\rceil$. 

Now, let $ d_n^{\gamma_n}(\cK)_X\asymp  [\log_2(n\varphi(n))]^\beta [n\varphi(n)]^{-\alpha}$.
If there is a constant $c_1>1$ for which the quantity $\displaystyle{\sup_{n\in \mathbb N} \frac{\varphi(c_1n)}{\varphi(n)}<\infty}$, then    
Lemma \ref{calculus}  (see the Appendix) gives that  $\displaystyle{\sup_{n\in \mathbb N} \frac{\varphi(c_0n)}{\varphi(n)}}$ is finite for  all $c_0>1$
and  the conclusion follows from  Lemma \ref{auxlemma}  (see the Appendix)  with $c=8$ and $a_k=\epsilon_k(\mathcal K)_X$. If for all 
 $c_0>1$ we have 
$\displaystyle{\sup_{n\in \mathbb N} \frac{\varphi(c_0n)}{\varphi(n)}=\infty}$, then we apply again Lemma \ref{auxlemma} to complete the proof.
 \hfill $\Box$

\bigskip
\begin{remark}
All functions $\varphi$ from Table {\rm 1} or Table {\rm 2} satisfy the condition
$\displaystyle{\sup_{n\in \mathbb N} \frac{\varphi(c_1n)}{\varphi(n)}<\infty}$ for all $c_1>1$,
and therefore it follows from Lemma {\rm \ref{newlemma}} that for $\gamma_n=2^{\varphi(n)}$ with $\varphi$ being any of these functions,
$$
d_n^{\gamma_n}(\cK)_X\asymp  [\log_2(n\varphi(n))]^\beta [n\varphi(n)]^{-\alpha}, \,n\in\mathbb N, \,\alpha>0,\,\beta\in\mathbb R\quad \Rightarrow\quad \epsilon_{n}(\cK)_X\asymp  [\log_2 n]^\beta n^{-\alpha}, \, n\in \mathbb N.
$$
\end{remark}

Observe that Lemma \ref{newlemma} does not cover  the case when $\gamma_n=const$, that is, $\varphi(n)=const$,  because the Lipschitz width 
$d_n^{\gamma}(\cK)_X\to 0$ as $n\to \infty$, and therefore would not satisfy the condition $d_n^{\gamma}(\cK)_X \geq c_02^{-\varphi(n)}=C$. We discuss this case separately in the lemma that follows.

\begin{lemma} 
\label{LL1new}
Let  $\cK \subset X$ be a compact set, $\gamma=const$, $(\xi_n)$ be a sequence of positive numbers such that $\xi_n\to 0$ as $n\to\infty$, and $d_n^{\gamma}(\cK)_X\geq \xi_n$.
Then we have that for some  positive constants $c,c_1$, 
\begin{equation}\label{(1)4}
d_n^{c_1{\xi_n^{-c}}}(\cK)_X\leq  \epsilon_{c\lceil n\log_2(\xi_n^{-1})\rceil}(\cK)_X\leq 3 d_n^\gamma (\cK)_X.
\end{equation}
In particular, if  $\xi_n=2^{-n}$, we have
$ d_n^{c_1 2^{cn}}(\cK)_X\leq \epsilon_{cn^2}(\cK)_X\leq 3d_n^\gamma (\cK)_X$,
and when $\xi_n=n^{-\alpha}$ we have
$  d_n^{c_1n^{c\alpha}}(\cK)_X\leq \epsilon_{c\lceil n\log_2 n\rceil}(\cK)_X\leq 3 d_n^\gamma (\cK)_X$.
\end{lemma}
\noindent
{\bf Proof:} Again, it follows from  the proof of Lemma \ref{LL1} that
for   $\epsilon=3d^{\gamma}_n(\cK)_X\geq 3\xi_n>0$  the entropy numbers $\epsilon_r(\cK)_X\leq 3d^{\gamma}_n(\cK)_X$, with
\be
\nonumber
r=\lceil n(\log_2\gamma+\log_2(2[d^{\gamma}_n(\cK)_X]^{-1}))\rceil\leq  c\lceil n\log_2 (\xi_n^{-1})\rceil,
\ee
if $n$ is large enough, and therefore
\begin{equation}
\label{nm}
\epsilon_{c\lceil n\log (\xi_n^{-1})\rceil}(\cK)_X\leq 3 d_n^\gamma (\cK)_X.
\end{equation}
We take $k_n$ to be the smallest integer such that $c\lceil n\log (\xi_n^{-1})\rceil \leq n k_n.$
From \cite[Th. 3.3]{PW}, (\ref{nm}), and the monotonicity of entropy numbers and Lipschitz widths we get 
$$
d_n^{c_1{\xi_n^{-c}}}(\cK)_X\leq  d_n^{2^{k_n} \mathrm{rad}(\cK)}(\cK)_X\leq \epsilon_{nk_n}(\cK)_X\leq \epsilon_{c\lceil n\log_2(\xi_n^{-1})\rceil}(\cK)_X\leq 3 d_n^\gamma (\cK)_X,
$$
 where we have used in the first inequality the definition of $k_n$, namely that $c\lceil n\log (\xi_n^{-1})\rceil > n (k_n-1)$.
The proof is completed. \hfill $\Box$

\section{Conclusion}
\label{con}
 In this paper, we further develop the theory of Lipschitz widths  for a compact set $\mathcal K$ in a Banach space $X$ to include Lipschitz mappings with large Lipschitz constants. 
The theory  is  then applied to NNs  to obtain Carl's type inequalities for deep and shallow neural network approximation,
 where the  error is measured in the Banach space norm $\|\cdot\|_X$ and the  requirement for $X$ is that   $C([0,1]^d)\subset X$ is continuously  embedded in $X$. 
In fact, this method can be used to obtain Carl's type inequalities 
for any feed-forward NN with predetermined 
relation between the width $W$ and depth $n$ of this network. 

Our analysis is executed by utilizing the growth of the  $\ell_\infty^m$ norm of the parameters of the NN, namely 
$\|y\|_{\ell_\infty^m }\leq w(m)$ (with $m=\tilde n$ or $m=\widetilde W$)  for a given function $w$.
Note that  all 
results for NN approximation utilize the behavior of the  Lipschitz width $d_m^{\gamma_m}(\mathcal K)_X$, which is defined as the infimum of the fixed Lipschitz width 
$d^{\gamma_m}(\mathcal K, Y_m)$ over all possible norms  $Y_m$ in $\mathbb R^m$. Therefore, all results will hold no matter what norm we choose to 
bound the NN parameters, i.e. all statements will hold if instead of $\|y\|_{\ell_\infty^{m} }\leq w(m)$, we choose $\|y\|_{Y_{m} }\leq w(m)$, where $Y_{m}$ is our favorite norm.

Carl's type inequalities for shallow NN approximation with certain activation functions (among  them 
 ${\rm ReLU}^k$) 
have been derived in \cite{JS} for some special compact sets $\mathcal K$. Our results for shallow neural network  approximation can be viewed as a generalization of the estimates 
in \cite{JS}.  In the case of deep NNs, lower bounds for DNN approximation have been derived for compact sets 
 $\mathcal K$ being the unit ball in $C^k$ or in certain Besov spaces, see \cite{DHP}, \S 5.9 and the references therein. The approach proposed there  uses the VC dimension of DNNs and the structure 
of the set $\mathcal K$. Compared to the lower estimates in \cite{DHP} in these special cases, 
our  results are inferior. This is expected since in our analysis we do  not utilize the particular structure of the class $\mathcal K$ but  only  its entropy numbers. As a result, our estimates can be applied to any novel classes 
as long as we know their entropy numbers.

\section{Appendix}
\label{ap}
In this section we provide the proofs of some of the theorems and lemmas we use in the paper.

 {\bf Proof of Theorem \ref{widthsfrombelownew}:}
We prove  the theorem  by  contradiction and start with (i). 
 If (\ref{widths(i)gen}) does not hold for some constant $C_1$,  then
there exists a strictly  increasing sequence of  natural numbers  $(n_k)_{k=1}^\infty$, such that
$$ 
p_k:= \frac{d_{n_k}^{\gamma_{n_k}}({\mathcal K})_X[n_k\varphi(n_k)]^{\alpha}}{[\log_2 (n_k\varphi(n_k))]^{\beta}}\to 0\quad\hbox{as}\quad k\to \infty.
$$
To simplify the notation, we denote $P_k:=n_k \varphi(n_k)$ and we then  write
\begin{equation} 
\nonumber
d_{n_k}^{\gamma_{n_k}} ({\mathcal K}) _X= \frac{p_k\left[\log_2P_k\right]^{\beta}}{P_k^{\alpha}}<
 \frac{2p_k\left[\log_2 P_k\right]^{\beta}}{P_k^{\alpha}}=:\delta_k, \quad  k=1,2,\dots.
\end{equation} 
We also denote $Q_k$ to be  $Q_k:= \log_2(3\gamma_{n_k}\delta_k^{-1})$ and  apply Proposition \ref{carl2}  with $\eta_n=c_1(\log_2n)^\beta n^{-\alpha}$
 and $\delta=\delta_k$ 
to obtain
$$
c_1 \left[\log_2 ({ n_k}Q_k)\right ]^\beta  {n_k}^{-\alpha}Q_k^{-\alpha } \leq 4 \frac{
p_k\left[\log_2 P_k\right]^{\beta}}{ [P_k]^{\alpha}}.
$$
We rewrite the latter inequality as
\begin{equation}
\label{ss}
p_k^{-1}\left[\frac{\log_2  (n_kQ_k)}{\log_2P_k}\right]^\beta  
\leq \frac{4}{c_1}  \left[\frac{Q_k}{\varphi(n_k)}\right]^{\alpha}.
\end{equation}
In what follows, we denote by $C$ a generic constant whose value may change every time.
Observe that 
\begin{eqnarray}  
\nonumber
Q_k&=&
c+\varphi(n_k) +\log_2 (p_k^{-1}) +\alpha\log_2 P_k-\beta\log_2(\log_2 P_k)\\
 \nonumber
 &\asymp&\varphi(n_k) +\log_2 (p_k^{-1} )+\log_2P_k\asymp\varphi(n_k) +\log_2 (p_k^{-1} )+\log_2n_k.
\nonumber
\end{eqnarray}
 Since for all $k$'s we have $\log_2n_k\leq c\varphi(n_k)$, then  
\begin{equation}
\label{ss6}
C\varphi(n_k)\leq Q_k 
\leq  C(\varphi(n_k) +\log_2 (p_k^{-1} )),
\end{equation}
and therefore
\begin{equation}
\label{ss7}
 \left[\frac{Q_k}{\varphi(n_k)}\right]^{\alpha}\leq 
 C\left[\frac{\varphi(n_k) +\log_2 (p_k^{-1} )}{\varphi(n_k)}\right]^{\alpha}\leq C
 \left[1+\log_2 (p_k^{-1} )\right]^{\alpha}.
\end{equation}
We now consider the following cases.

\noindent
\underline {\bf Case 1: $\beta\geq 0$}. { It follows from (\ref{ss6}) that}
 $$
 \frac{\log_2  (n_kQ_k)}{\log_2 P_k}\geq {C}  \frac{\log_2 P_k}{\log_2 P_k}= {C},
 $$
 { which combined with  (\ref{ss})  and (\ref{ss7})  gives}
 $$
 p_k^{-1}\leq  { \frac{4}{c_1}\left[\frac{\log_2  (n_kQ_k)}{\log_2P_k}\right]^{-\beta}  
  \left[\frac{Q_k}{\varphi(n_k)}\right]^{\alpha}}\leq C\left[1+\log_2 (p_k^{-1} )\right]^{\alpha},
 $$
 which contradicts the fact that $p_k^{-1}\to \infty$.

\noindent
\underline {\bf Case 2: $\beta<0$.}   It follows from (\ref{ss6}) that
\begin{equation}
\label{ss8}
\frac{\log_2  ({n_k}Q_k)}{\log_2P_k}
\leq C\frac{\log_2  ({n_k}(\varphi(n_k) +\log_2 (p_k^{-1} )))}{\log_2 P_k},
\end{equation}
and we consider two cases.

\underline {\bf Case 2.1:}  If for infinitely many $k$'s we have $\log_2 (p_k^{-1} )\leq \varphi(n_k)$, then for those $k$'s  (\ref{ss8}) 
becomes
$$
\frac{\log_2  ({n_k}Q_k))}{\log_2 P_k}
\leq C\frac{\log_2  ({n_k}(2\varphi(n_k)))}{\log_2 P_k}\leq C,
$$
and {   (\ref{ss}) gives} 
\begin{eqnarray*}
p_k^{-1}&\leq&  \frac{4}{c_1}\left[\frac{\log_2  ({n_k}Q_k))}{\log_2P_k}\right]^{-\beta}  
\left[1+\log_2 (p_k^{-1} )\right]^{\alpha}\leq C\left[1+\log_2 (p_k^{-1} )\right]^{\alpha},
\end{eqnarray*}
which contradicts the fact that $p_k^{-1}\to \infty$.

\underline {\bf Case 2.2:}  If for infinitely many $k$'s we have $\log_2 (p_k^{-1} )\geq \varphi(n_k)\geq  c\log_2n_k$, 
then (\ref{ss8}) 
becomes
\begin{eqnarray*}
\frac{\log_2  ({n_k}Q_k)}{\log_2 P_k}
&\leq& C\log_2  (n_k\log_2(p_k^{-1}))
=C(\log_2  n_k+\log_2(\log_2(p_k^{-1})))\\
&\leq &C(\log_2(p_k^{-1})+\log_2  (\log_2(p_k^{-1}))){\leq C\log_2(p_k^{-1})},
\end{eqnarray*}
and leads to
\begin{eqnarray*}
p_k^{-1}&\leq& \frac{4}{c_1}\left[\frac{\log_2  ({n_k}Q_k))}{\log_2 P_k}\right]^{-\beta}  
\left[1+\log_2 (p_k^{-1} )\right]^{\alpha}
   \leq C\left[\log_2(p_k^{-1})\right]^{-\beta}\left[1+\log_2 (p_k^{-1} )\right]^{\alpha},
\end{eqnarray*}
which contradicts the fact that $p_k^{-1}\to \infty$.

To prove (ii), we repeat the argument for (i), namely, we assume that (ii) does not hold, and therefore, there exists a strictly  increasing sequence of natural numbers 
$(n_k)_{k=1}^\infty$, such that
$$ 
e_k:= d_{n_k}^{\gamma_n}({\mathcal K})_X[D_k]^{\alpha}\to 0\quad\hbox{as}\quad k\to \infty,
$$
where $D_k:=\log_2 (n_k\varphi(n_k))$. 
We write
\begin{equation} 
\label{widths(iii)new}
d_{n_k}^{ \gamma_{n_k}} ({\mathcal K})_X= e_k[D_k]^{-\alpha}<
2e_k[D_k]^{-\alpha}=:\delta_k, \quad k=1,2,\dots,
\end{equation} 
and use  Proposition \ref{carl2}  with $\eta_n=c_1(\log_2n)^{-\alpha}$
to derive
$$
c_1 \left[\log_2 (n_k Q_k  )\right ]^{-\alpha}
 \leq 4e_k D_k^{-\alpha},
$$
where as in (1) we have the notation $Q_k:=\log_2(3\gamma_{n_k}\delta_{k}^{-1})$. We rewrite this inequality as
\begin{equation} 
\label{f111}
e_k^{-1}\leq \frac{4}{c_1}\left[\frac{\log_2 (n_k Q_k)}{D_k}\right ]^{\alpha}.
\end{equation}
Since $\log_2n_k\leq c\varphi(n_k)$, we have 
\begin{eqnarray*}
Q_k&=&\log_2(1.5)+\varphi(n_k)+\log_2(e_k^{-1})+\alpha\log_2D_k\\
&\leq&C[\varphi(n_k)+\log_2(e_k^{-1})+\log_2(\log_2n_k+\log_2(\varphi(n_k)))]\\
&\leq&C[\varphi(n_k)+\log_2(e_k^{-1})+\log_2(\varphi(n_k)+\log_2(\varphi(n_k)))]\leq C[\varphi(n_k)+\log_2(e_k^{-1})],
\end{eqnarray*}
and (\ref{f111}) becomes
\begin{equation}
\label{qq}
e_k^{-1}\leq C\left[\frac{\log_2 (n_k(\varphi(n_k)+\log_2(e_k^{-1})))}{D_k}\right]^{\alpha }.
\end{equation}

\noindent
\underline{\bf Case 1:}   If for infinitely many values of $k$ we have $\log_2(e_k^{-1})\leq \varphi(n_k)$, then 
$$
e_k^{-1}\leq C,
$$
which contradicts with the fact that $e_k^{-1}\to \infty$ as $k\to \infty$.

\noindent
\underline{\bf Case 2:}   If for infinitely many values of $k$ we have $\log_2(e_k^{-1})\geq \varphi(n_k)\geq c\log_2n_k$, then for those $k$'s  we get $D_k \geq {C}$,  
so (\ref{qq})  becomes
\begin{eqnarray*}
e_k^{-1}&\leq& C\left[{\log_2 (n_k(\varphi(n_k)+\log_2(e_k^{-1})))}\right]^{\alpha }=
 C\left[\log_2 n_k+\log_2(\varphi(n_k)+\log_2(e_k^{-1}))\right]^{\alpha }\\
 &\leq& 
C\left[\log_2 (e_k^{-1})+\log_2(\log_2(e_k^{-1}))\right]^{\alpha}\leq C\left[\log_2 (e_k^{-1})\right]^{\alpha},
\end{eqnarray*}
which also contradicts with the fact that $e_k^{-1}\to \infty$ as $k\to \infty$,
and the proof is completed.
\hfill $\Box$

\bigskip
\noindent
{\bf Proof of Lemm \ref{LL1}:} It follows from Proposition 3.6 in \cite{PW}  that if $d^{\gamma_n}_n(\cK)_X< \epsilon/2$, then 
$$
\gamma_n\geq\frac{1}{6}\epsilon N_{\epsilon}^{1/n}(\mathcal K)\quad\Rightarrow\quad N_{\epsilon}(\cK)\leq\left[\frac{6\cdot 2^{\varphi(n)}}{\epsilon}\right]^n\leq 2^r,
$$
where $N_\epsilon(\mathcal K)$ is the $\epsilon$ covering number of $\mathcal K$ (the cardinality of the minimal $\epsilon$-covering of $\mathcal K$)
and 
$$
r=r(n) =\lceil n(\varphi(n)+ \log_2(6/\epsilon))\rceil
$$
 is the smallest integer that this inequality holds.  We rewrite the above as 
$\epsilon_r(\cK)_X\leq \epsilon$.

If $d_n^{\gamma_n}(\cK)_X=0$, 
then for any  $k\in \mathbb N$    such that    $k>\varphi(n)$, we take $\epsilon=6\cdot 2^{\varphi(n)-k}$  and  obtain $r=nk$ and
$$
\epsilon_{nk}(\mathcal K)_X\leq 6\cdot 2^{\varphi(n)-k}= 6\cdot \gamma_n 2^{-k}.
$$
This is an optimal estimate since for the ball $\cK_n:=\{y\in \mathbb R^n\ :\ \|y\|_{\ell_2^n} \leq \gamma_n\}$  of radius $\gamma_n$ in   $\mathbb R^n$,  we have that
$d_n^{\gamma_n}(\cK_n)_X=0$ since for the $\gamma_n$-Lipschitz mapping $\Phi(y)=\gamma_ny$ we have $\cK_n=\Phi(B_{\ell_2^n}(1))$. On the other hand,   it is a well known fact that for any integer $s$ we have, see \cite{CS}, 
$$  \gamma_n 2^{-s/n}\leq \epsilon_s(\cK_n)_X\leq 4\gamma_n 2^{-s/n}.$$

Now we continue with  the other cases of the behavior of the Lipschitz width $d^{\gamma_n}_n(\cK)_X$  under the assumption that $\gamma_n= 2^{cn^p[\log_2n]^q}$.
In the case when  $\epsilon=3d^{\gamma_n}_n(\cK)_X$ with $d^{\gamma_n}_n(\cK)_X>0$, we obtain that $\epsilon_r(\cK)_X\leq 3d^{\gamma_n}_n(\cK)_X$, where 
\begin{equation}
\label{wes}
r(n)=\lceil n(\varphi(n)+\log_2(2[d^{\gamma_n}_n(\cK)_X]^{-1}))\rceil\leq \lceil n(2\varphi(n)-\log_2 (d^{\gamma_n}_n(\cK)_X))\rceil.
\end{equation}

In the case when $0<d^{\gamma_n}_n(\cK)_X\lesssim [\log_2n]^\beta n^{-\alpha}$, it follows from \eref{wes} that 
\begin{eqnarray*}
r(n)
\lesssim \lceil n^{p+1}[\log_2 n]^q+ n\alpha \log_2 n -n \beta\log_2(\log_2n)\rceil\lesssim  \lceil n^{p+1}[\log_2 n]^q+ n\log_2 n\rceil
\end{eqnarray*}
provided  $n$ is sufficiently large.

\underline{\bf Case 1:  $p> 0$ and $q\in \mathbb R$ or $p=0$ and $q\geq 1$}. For this range of $p$ and $q$ we  have 
$$
r(n)\lesssim n^{1+p}[\log_2 n]^q.
$$
In this case, for  each $k\in \mathbb N$  large enough, we consider   $n=n(k)$, defined as
$$
n=\lceil k^{1/(p+1)}[\log_2 k]^{-q/(p+1)}\rceil<2k^{1/(p+1)}[\log_2 k]^{-q/(p+1)},
$$
 and for $k$ large enough we have
\begin{eqnarray*}
r(n)&\lesssim&\left[ k^{1/(p+1)}[\log_2 k]^{-q/(p+1)} \right]^{p+1}[\log_2 (k^{1/(p+1)}[\log_2 k]^{-q/(p+1)})]^q\\
&=& k [\log_2 k]^{-q}\left[ (p+1)^{-1}\log_2 k-\frac{q}{p+1}\log_2( \log_2 k)\right]^q.
\end{eqnarray*}
Note that
\begin{equation}
\label{ds}(p+1)^{-1}\log_2 k-\frac{q}{p+1}\log_2 \log_2 k\asymp \log_2 k,
\end{equation}
and therefore 
$$
r(n)\lesssim
k [\log_2 k]^{-q}\left[ \log_2 k\right]^q
= k 
$$
Since $\epsilon_k(\cK)_X$ is a monotone decreasing sequence, we have that
\begin{eqnarray*}
\epsilon_{ck(}\cK)_X&\leq& \epsilon_{r(n)}(\cK)_X\leq 3d^{\gamma_n}_n(\cK)_X\lesssim [\log_2n]^\beta n^{-\alpha}\\
&\lesssim&[\log_2 ( k^{1/(p+1)}[\log_2 k]^{-q/(p+1)})]^\beta [k^{1/(p+1)}[\log_2 k]^{-q/(p+1)}]^{-\alpha}\\
&=&\left[\frac{1}{p+1}\log_2 k -\frac{q}{p+1}\log_2( \log_2 k)\right]^\beta k^{-\alpha/(p+1)} 
[\log_2 k]^{q\alpha/(p+1)}
\end{eqnarray*}
It follows from \eref{ds} that
$$
\epsilon_{ck(}\cK)_X\lesssim
\left[\log_2 k\right]^\beta k^{-\alpha/(p+1)} 
[\log_2 k]^{q\alpha/(p+1)}
= k^{-\alpha/(p+1)}\left[ \log_2 k  \right]^{\beta+\frac{\alpha q}{p+1}}.
$$

\underline{\bf Case 2:   $p=0$ and $q< 1$}. For this range of $p$ and $q$ we  have $r(n)\lesssim n\log_2 n$.
In this case, for  each $k\in \mathbb N$ large enough we consider  $n=n(k)$, given as
$$
n=\lceil k[\log_2 k]^{-1}\rceil<2k[\log_2 k]^{-1},
$$
 and for $k$ large enough we have
\begin{eqnarray*}
r(n)\lesssim k[\log_2 k]^{-1} \log_2 (k[\log_2 k]^{-1})= k [\log_2 k]^{-1}\left[ \log_2 k-\log_2 (\log_2 k)\right].
\end{eqnarray*}
Using the fact that 
 $\log_2 k-\log_2 (\log_2 k)\asymp \log_2 k$,
we conclude that 
$
r(n)\lesssim k,
$
and as in Case 1, we have that
\begin{eqnarray*}
\epsilon_{ck(}\cK)_X&\leq& \epsilon_{r(n)}(\cK)_X\leq 3d^{\gamma_n}_n(\cK)_X\lesssim [\log_2n]^\beta n^{-\alpha}
\lesssim [\log_2 ( k[\log_2 k]^{-1})]^\beta [k[\log_2 k]^{-1}]^{-\alpha}\\
&=&\left[\log_2 k -\log_2 (\log_2 k)\right]^\beta k^{-\alpha}
[\log_2 k]^{\alpha} \lesssim k^{-\alpha}[\log_2 k]^{\beta+\alpha},
\end{eqnarray*}
and the proof for this particular case is completed.

Now,  when $0<d^{\gamma_n}_n(\cK)_X\lesssim 2^{-cn}$, it follows from \eref{wes} that 
\begin{eqnarray*}
r(n)
\lesssim  n(\varphi(n)+n)\lesssim \lceil n^{p+1}[\log_2n]^q+n^2\rceil.
\end{eqnarray*}

\underline{\bf Case 1:   $0\leq p<1$ and  $q\in \mathbb R$  or $p=1$ and  $q\leq 0$}. For this range of $p$ and $q$ we  have 
$$
r(n)\lesssim n^2.
$$
In this case, for  each $k\in \mathbb N$  large enough  we consider   $n=n(k)$ given by
$$
n=\lceil \sqrt{k}\rceil \leq 2\sqrt{k},
$$
 and for $k$ large enough we have $r(n)\lesssim k$ and thus
\begin{eqnarray*}
\epsilon_{\tilde ck(}\cK)_X&\leq& \epsilon_{r(n)}(\cK)_X\leq 3d^{\gamma_n}_n(\cK)_X\lesssim 2^{-cn}\lesssim 2^{-c\sqrt{k}}.
\end{eqnarray*}

\underline{\bf Case 2:   $p>1$ and  $q\in \mathbb R$  or $p=1$ and  $q>0$}. In this case 
$$
r(n)\lesssim n^{p+1}[\log_2n]^q,
$$
and  for  each $k\in \mathbb N$  large enough, we define $n=n(k)$ via  
$$
n=\lceil k^{1/(p+1)}[\log_2 k]^{-q/(p+1)}\rceil<2k^{1/(p+1)}[\log_2 k]^{-q/(p+1)},
$$
 and for $k$ large enough, as before,  we have
\begin{eqnarray*}
r(n)&\lesssim&\left[ k^{1/(p+1)}[\log_2 k]^{-q/(p+1)} \right]^{p+1}[\log_2 (k^{1/(p+1)}[\log_2 k]^{-q/(p+1)})]^q\\
&=& k [\log_2 k]^{-q}\left[ (p+1)^{-1}\log_2 k-\frac{q}{p+1}\log_2( \log_2 k)\right]^q\lesssim
k [\log_2 k]^{-q}\left[ \log_2 k\right]^q
= k.
\end{eqnarray*}
Therefore, we obtain 
$$
\epsilon_{\tilde ck(}\cK)_X\lesssim d^{\gamma_n}_n(\cK)_X\lesssim 2^{-cn}\leq 2^{-{c}[k[\log_2k]^{-q}]^{1/(p+1)}}.
$$
This completes the proof of the lemma.
\hfill $\Box$

We now state and proof a lemma that we used in \S\ref{Apclasses}  and \S\ref{ACSSN}.

\begin{lemma}
\label{Pr}
Let $\Sigma_n$, $n=1,2\dots$, be compact  subsets of a Banach space $Y$ which is continuously embedded in the Banach space $X$ and let $(\xi_n)_{n=1}^\infty$ be a sequence of non-negative numbers such that $\inf_n\xi_n=0$ ( in particular $\lim_{n\to \infty}\xi_n=0$).  If  $V_n:=\{f\in X \ :\ {\rm dist} (f,\Sigma_n)_X\leq \xi_n\}$, then  the set $\cK:=\bigcap_{n=1}^\infty V_n$ is compact in $X$.
\end{lemma}
\noindent
{\bf Proof:} Since  each $\Sigma_n$ is a compact set,  each $V_n$ is a closed and bounded subset of $X$. This implies that $\cK$ is closed and bounded (possibly empty) subset of $X$.
 To prove that $\cK$ is compact, we argue via contradiction. Let us assume that there exists $\delta>0$ and a sequence of elements $(f_j)_{j=1}^\infty \subset \cK$ such that $\|f_j-f_{j^\prime}\|_X \geq \delta$ whenever $j\neq j^\prime$. Let us fix  $n$ such that $\xi_n\leq \delta/3$. Since $(f_j)_{j=1}^\infty \subset V_n$, for $j\in \NN$ we have $f_j=v_j +q_j$, where $v_j\in \Sigma_n$ and $\|q_j\|_X \leq \xi_n$. Thus
$$
\|v_j-v_{j^\prime}\|_X=\|f_j-f_{j'}-q_j+q_{j'}\|_X\geq \|f_j-f_{j'}\|_X-\|q_j\|_X-\|q_{j'}\|_X\geq \delta-2\xi_n\geq \delta/3,
$$ 
which means that 
$$
\delta/3\leq \|v_j-v_{j^\prime}\|_X\leq C\|v_j-v_{j^\prime}\|_Y,
$$
and thus $\Sigma_n$ is {\em not compact.} The proof is completed.
\hfill $\Box$

\bigskip

 We next state  a simple lemma that we  use in the proof of Lemma \ref{auxlemma}.
\begin{lemma}
\label{calculus}
The following holds
\begin{itemize}
\item The function $f_{\alpha,\beta}:[1,\infty)\to\mathbb R$, defined as $f_{\alpha,\beta}(t):=t^{-\alpha}[\log_2 t]^\beta$, for $\alpha>0$, $\beta\in \mathbb R$  is bounded,   monotonically decreasing on $[1,\infty)$ when $\beta\leq 0$ and on $[2^{\beta/\alpha},\infty)$ when $\beta\geq 0$.
 \item If $\varphi:[1,\infty)\to \mathbb R$ is an increasing non-negative function then 
 $$
 \exists \,\,c_1>1, \,\,\hbox{such that }\,\,\sup_{n\in \mathbb N}\frac{\varphi(c_1n)}{\varphi(n)}<\infty\quad \Leftrightarrow\quad \forall\,\, c>1,\,\sup_{n\in \mathbb N}\frac{\varphi(cn)}{\varphi(n)}<\infty.
 $$
\end{itemize}
\end{lemma}
\noindent
{\bf Proof:} The proof is simple calculus and we omit it.
\hfill$\Box$

\begin{lemma} 
\label{auxlemma}
Let $(a_n)$ be a monotone non-increasing sequence and  
  $\varphi:[1,\infty)\to \mathbb R$ be an increasing non-negative function such that $\varphi(n)\to\infty$ as $n\to\infty$. If there is a constant $c>1$ such that 
  \begin{equation}
  \label{res31new} 
A_n:=\frac{ [\log_2(cn\varphi(cn))]^\beta}{ [cn\varphi(cn)]^{\alpha}}\lesssim a_{c n\lceil\frac{1}{2}\varphi(cn)\rceil}\lesssim\frac{ [\log_2(n\varphi(n))]^\beta }{[n\varphi(n)]^{\alpha}}=:B_n,
\end{equation}
then
\begin{itemize}
\item  if  the function $\varphi$ is such that there is a constant $c_1>1$ for which $\displaystyle{\sup_{n\in \mathbb N} \frac{\varphi(c_1n)}{\varphi(n)}<\infty}$ then  
$$
a_{m}\asymp  (\log_2 m)^\beta m^{-\alpha}, \quad m=1,2,\ldots.
$$
\item { if there is $c>1$ such that    $\displaystyle{\sup_{n\in \mathbb N} \frac{\varphi(cn)}{\varphi(n)}= \infty}$, then } the lower and upper bound in  {\rm (\ref{res31new})} are asymptotically different in the sense that
$$
 \frac{B_n}{A_n}\gtrsim \begin{cases}\left[\frac{\varphi(cn)}{\varphi(n)}\right]^\alpha
\left[ \log_2 \frac{\varphi(cn)}{\varphi(n)}\right]^{-\beta}, \quad \beta>0,\\ \\
 \left[\frac{\varphi(cn)}{\varphi(n)}\right]^\alpha, \quad \beta\leq 0.
 \end{cases}
$$
\end{itemize}
\end{lemma}
\noindent
{\bf Proof:} 
It follows from Lemma \ref{calculus} that the quantity  $\displaystyle{\sup_{n\in \mathbb N} \frac{\varphi(c_1n)}{\varphi(n)}}$ can be either finite for some $c_1>1$ (and therefore for all $c>1$) or infinite for    every 
{ $c_1>1$}.  
Let us first consider the former case, which,  according to the same lemma, gives that
$\displaystyle{\sup_{n\in \mathbb N} \frac{\varphi(cn)}{\varphi(n)}:=\rho=\rho(c)<\infty}$. We next bound  the quantity $ \frac{B_n}{A_n}$ as
$$
\frac{B_n}{A_n}={c^\alpha}\left[\frac{\log_2 (n\varphi(n))}{\log_2(cn\varphi(cn))}  \right]^\beta \left[ \frac{\varphi(cn)}{\varphi(n)} \right]^\alpha
=:{c^\alpha}R_n^\beta S_n^\alpha\leq \tilde C,
$$
since  $S_n\leq \rho$ and 
$$
1\leq R_n^{-1}=\frac{\log_2 (cn\varphi(cn))}{\log_2(n\varphi(n))} \leq \frac{\log_2(c  \rho)+\log_2 (n\varphi(n))} {\log_2 (n \varphi(n))} =1+\frac{\log_2(c  \rho)} {\log_2 n \varphi(n)} \leq \tilde c,
$$
for $n$ large enough.
Thus, (\ref{res31new}) becomes
\begin{equation}
\label{mm}
A_n\lesssim a_{c n\lceil\frac{1}{2}\varphi(cn)\rceil}\lesssim A_n.
\end{equation}
We now fix $k$ and choose $n=n(k)$ such that 
$$
c n\lceil\frac{1}{2}\varphi(cn)\rceil<k\leq c (n+1)\lceil\frac{1}{2}\varphi(c(n+1))\rceil.
$$
Clearly, because of the monotonicity of $(a_k)$ and (\ref{mm}), for such $k$ we have
\begin{equation}
\label{nn}
A_{n+1}\lesssim a_k\lesssim A_n.
\end{equation}
Next, let us consider the ratio
\begin{eqnarray}
\nonumber
\frac{A_n}{A_{n+1}}=\left[ \frac{\log_2(cn\varphi(cn))}{\log_2(c(n+1)\varphi(c(n+1)))}\right]^\beta\left[ \left(1+\frac{1}{n}\right)\frac{\varphi(c(n+1))}{\varphi(cn)}\right]^\alpha=:P_n^\beta Q_n^\alpha.
\end{eqnarray}
Note that we have 
$$
\frac{A_n}{A_{n+1}}\leq C_1.
$$
Indeed, since
$$
\frac{\varphi(c(n+1))}{\varphi(cn)}\leq \frac{\varphi(2cn)}{\varphi(cn)} \leq \frac{\varphi(2cn)}{\varphi(n)}=
\frac{\varphi(2cn)}{\varphi(2n)}\frac{\varphi(2n)}{\varphi(n)}\leq
\rho(c)\rho(2),
$$
we have 
$$
Q_n= \left(1+\frac{1}{n}\right)\frac{\varphi(c(n+1))}{\varphi(cn)}\leq 
\frac{3}{2}\rho(c)\rho(2),
$$
and 
\begin{eqnarray}
\nonumber
1&\leq&  P_n^{-1}=\frac{\log_2(c(n+1)\varphi(c(n+1)))}{\log_2(cn\varphi(cn))}\leq \frac{\log_2(2cn\rho(c)\rho(2)\varphi(cn))}{\log_2(cn)+\log_2(\varphi(cn))}\\
\nonumber
&=&\frac{1+\log_2(cn)+\log_2(\varphi(cn))+\log_2(\rho(c)\rho(2))}{\log_2(cn)+\log_2\varphi(cn)}=1+\frac{\log_2(\rho(c)\rho(2))}{\log_2(cn)+\log_2\varphi(cn)}\leq 2,
\end{eqnarray}
for $n$ large enough. Then, we can rewrite (\ref{nn}) as
$$
A_n\lesssim  a_k \lesssim  A_n, \quad \hbox{for}\quad c n\lceil\frac{1}{2}\varphi(cn)\rceil<k\leq c (n+1)\lceil\frac{1}{2}\varphi(c(n+1))\rceil.
$$
For any $k$ in this interval we have that 
$$
\frac{c}{2}n\varphi(cn)<k\leq c(n+1)\left (\frac{1}{2}\varphi(c(n+1))+1\right)<c(n+1)\varphi(c(n+1))
$$
for $n=n(k)$ big enough.
Therefore, it follows from Lemma \ref{calculus} that 
$$
f_{\alpha,\beta}(2k)\leq f_{\alpha,\beta}(cn\varphi(cn))=A_n\lesssim a_k\lesssim A_n\lesssim A_{n+1}{=f_{\alpha,\beta}(c(n+1)\varphi(c(n+1)))}\leq f_{\alpha,\beta}(k),
$$
which implies that $a_k\asymp k^{-\alpha} [\log_2k]^\beta$ with constants depending on $\rho$, $\alpha$ and $\beta$.

Finally, in the case when there is a $c>1$ such that $\displaystyle{\sup_{n\in \mathbb N}\frac{\varphi(cn)}{\varphi(n)}=\infty}$ (and therefore this holds for every $c>1$), we have that 
$$
\frac{B_n}{A_n}={c^{\alpha}}\left[\frac{\log_2 (n\varphi(n))}{\log_2 (cn\varphi(cn))}  \right]^\beta \left[ \frac{\varphi(cn)}{\varphi(n)} \right]^\alpha\geq {c^{\alpha}}\left[ \frac{\varphi(cn)}{\varphi(n)} \right]^\alpha, \quad \hbox{for}\quad \beta\leq 0.
$$
When $\beta>0$ we write $\varphi(cn)=k(n)\varphi(n)$ and have
\begin{eqnarray}
\nonumber
\frac{B_n}{A_n}= {c^{\alpha}}[k(n)]^\alpha  \left[\frac{\log_2(n\varphi(n))   } {\log_2c+\log_2(k(n))+\log_2(n\varphi(n))   }  \right]^\beta.
\end{eqnarray}
We consider $n$ large enough so that $n\varphi(n)\geq c$.

\underline {Case 1:} If $k(n)\geq n\varphi(n)$ we have that 
$$
\frac{B_n}{A_n}\gtrsim [\log_2(n\varphi(n))]^\beta \frac{ [k(n)]^\alpha}{[3\log_2(k(n))]^\beta} \gtrsim\frac{ k(n)^\alpha}{[\log_2(k(n))]^\beta},
$$
for $n$ large enough.

\underline {Case 2:}   If $k(n)<n\varphi(n)$ we obtain that
$$\frac{B_n}{A_n}\gtrsim [k(n)]^\alpha \left[  \frac{\log_2(n\varphi(n))   } {2\log_2(n\varphi(n)) +\log_2c }   \right]^\beta{\geq [k(n)]^\alpha3^{-\beta}}\gtrsim[k(n)]^\alpha,
$$
which concludes the proof of the lemma.
  \hfill $\Box$

We do not know whether in the case when ${\displaystyle\sup_{n\in \mathbb N}\frac{\varphi(cn)}{\varphi(n)}=\infty}$, the discrepancy of the behavior of $\frac{B_n}{A_n}$ 
for positive and negative $\beta$'s is supported by examples or is due to our approach.

\vskip .1in
\noindent
{\bf Affiliations:}

\noindent
Guergana Petrova, Department of Mathematics, Texas A$\&$M University, College Station, TX 77843,  gpetrova$@$math.tamu.edu
\vskip .1in
\noindent
Przemys{\l}aw Wojtaszczyk, Institut of Mathematics Polish Academy of Sciences, ul. 
{\'S}niadeckich 8,  00-656 Warszawa, Poland, wojtaszczyk$@$impan.pl

\end{document}